\pdfoutput=1

\documentclass[10pt,twocolumn,letterpaper]{article}

\usepackage{times}
\usepackage{xspace}
\usepackage[dvipsnames]{xcolor}
\usepackage{graphicx}
\makeatletter\let\fh@pdfximage\pdfximage\def\pdfximage{\immediate\fh@pdfximage}\makeatother
\usepackage{amsmath}
\usepackage{amssymb}
\usepackage{booktabs}
\usepackage[numbers,sort&compress]{natbib}
\usepackage[hyphens]{url}
\usepackage{etoolbox}
\usepackage[format=plain,labelformat=simple,labelsep=period,font=small,compatibility=false]{caption}
\usepackage[font=footnotesize,skip=3pt,subrefformat=parens]{subcaption}
\usepackage[shortlabels,inline]{enumitem}
\setlist[itemize]{noitemsep,leftmargin=*,topsep=0em}
\setlist[enumerate]{noitemsep,leftmargin=*,topsep=0em}
\makeatletter
\font\elvbf = ptmb scaled 1100
\font\tenbf = ptmb scaled 1000
\def\fhsection{\@startsection{section}{1}{\z@}{-10pt plus -2pt minus -2pt}{7pt}{\large\bf}}
\def\fhssect#1{\fhsection*{#1}}
\def\fhsect#1{\fhsection{\texorpdfstring{\hskip -1em.~}{}#1}}
\def\section{\@ifstar\fhssect\fhsect}
\def\fhsubsection{\@startsection{subsection}{2}{\z@}{-8pt plus -2pt minus -2pt}{5pt}{\elvbf}}
\def\fhssubsect#1{\fhsubsection*{#1}}
\def\fhsubsect#1{\fhsubsection{\texorpdfstring{\hskip -1em.~}{}#1}}
\def\subsection{\@ifstar\fhssubsect\fhsubsect}
\def\fhsubsubsection{\@startsection{subsubsection}{3}{\z@}{-6pt plus -2pt minus -2pt}{3pt}{\tenbf}}
\def\fhssubsubsect#1{\fhsubsubsection*{#1}}
\def\fhsubsubsect#1{\fhsubsubsection{\texorpdfstring{\hskip -1em.~}{}#1}}
\def\subsubsection{\@ifstar\fhssubsubsect\fhsubsubsect}

\def\@maketitle{%
  \newpage\null\vskip .1in
  \begin{center}
    {\Large\bf \@title \par}
    \vspace*{13pt}
    {\large\lineskip .5em
     \begin{tabular}[t]{c}\@author\end{tabular}\par}
    \vskip .5em\vspace*{12pt}
  \end{center}}
\def\abstract{\centerline{\large\bf Abstract}\vspace*{12pt}\noindent\it\ignorespaces}

\let\titleold\title
\renewcommand{\title}[1]{\titleold{#1}\newcommand{\thetitle}{#1}}

\DeclareRobustCommand\onedot{\futurelet\@let@token\@onedot}
\def\@onedot{\ifx\@let@token.\else.\null\fi\xspace}

\makeatother

\usepackage{microtype}

\makeatletter\newcommand\captionof[1]{\def\@captype{#1}\caption}\makeatother
\usepackage{xspace}
\AtBeginDocument{\crefname{table}{Table}{Tables}\crefname{figure}{Figure}{Figures}}
\newcommand{\suppsec}[2]{\ifcsname r@#1\endcsname\hyperref[#1]{Sec.~#2}\else Sec.~#2\fi}
\newcommand{\paper}{ForceHarness\xspace}
\newcommand{\bench}{HiddenPhys\xspace}
\newcommand{\tool}[1]{\textit{#1}}

\usepackage{multirow}
\usepackage{array}

\definecolor{linkblue}{rgb}{0.21,0.49,0.74}
\usepackage[breaklinks,colorlinks,allcolors=linkblue]{hyperref}
\usepackage[capitalize]{cleveref}
\crefname{section}{Sec.}{Secs.}
\Crefname{section}{Section}{Sections}
\Crefname{table}{Table}{Tables}
\crefname{table}{Tab.}{Tabs.}

\title{ForceHarness: Turning Force into Action and Physics\\for Frozen Frontier Models}
\newcommand{\projecturl}{https://showlab.github.io/ForceHarness}
\author{Ke Sun\quad Kevin Yuchen Ma\quad Yike Zhang\quad Yanzhe Chen\quad Mike Zheng Shou$^{\dagger}$\\[0.25em]
{\normalsize Show Lab, National University of Singapore}\\[0.25em]
{\small Project page: \href{\projecturl}{showlab.github.io/ForceHarness}}
}
\hypersetup{
  pdftitle={ForceHarness: Turning Force into Action and Physics for Frozen Frontier Models},
  pdfauthor={Ke Sun, Kevin Yuchen Ma, Yike Zhang, Yanzhe Chen, Mike Zheng Shou}}

\begin{document}
\addtocontents{toc}{\protect\setcounter{tocdepth}{0}}
\twocolumn[{%
\renewcommand\twocolumn[1][]{#1}%
\maketitle
\begin{center}
  \includegraphics[width=\textwidth]{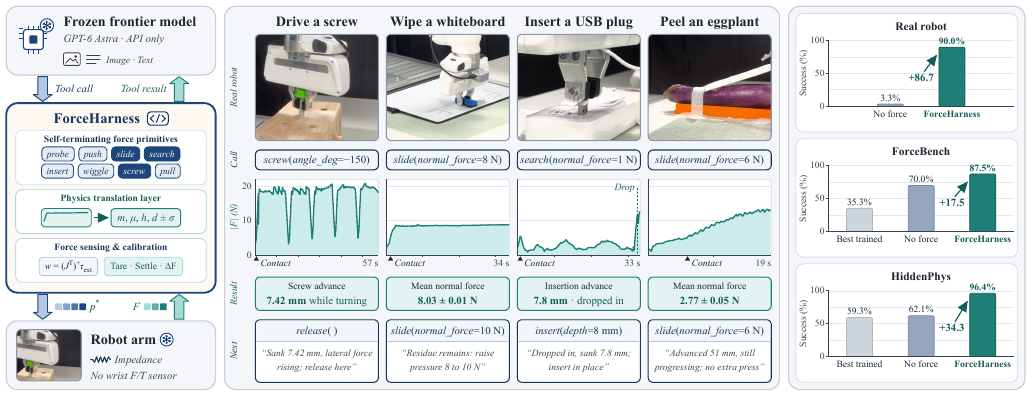}
  \captionof{figure}{\paper lets a frozen frontier model use wrist force.
  Left: the model only issues tool calls; the harness calibrates hand force, runs self-terminating force primitives beneath the model, and returns physical quantities.
  Middle: four real-robot tasks; each column shows one primitive call, the force logged during that call, the quantity its tool result reported, and the model's next call with a summary of its reason.
  Right: success rate of GPT-6 Astra with and without force on the real robot, ForceBench, and \bench, with the best trained policy where one exists.}
  \label{fig:teaser}
\end{center}%
\vspace{0.5em}
}]
{\renewcommand{\thefootnote}{\fnsymbol{footnote}}\footnotetext[2]{Corresponding author.}}
\begin{abstract}
Frontier multimodal models can now control robots zero-shot from images, yet many manipulation tasks depend on contact forces that images cannot show.
Bringing force to an API-only frozen model raises two challenges.
First, contact evolves much faster than an API model can decide, so the system must close the loop on force beneath the model.
Second, a frozen model cannot interpret raw force streams, so force must be returned in a form the model can reason about.
Therefore, we present \paper, a harness that lets frozen models use force without demonstrations or finetuning.
After calibrating hand force from a wrist sensor or joint torques alone, \paper addresses the first challenge with self-terminating force primitives that regulate force on their own and return at a contact event.
To address the second challenge, a physics translation layer turns each force trace into physical quantities with uncertainty as the tool result.
To evaluate \paper against expert force controllers and learned policies under identical conditions, we also build \bench, a seven-task MuJoCo benchmark with hidden physical variables.
On a real arm without a wrist force sensor, \paper raises the zero-shot success rate of GPT-6 Astra from 3.3\% to 90.0\%.
In simulation, it adds 17.5 percentage points on the public ForceBench suite and 34.3 percentage points on \bench, surpassing force-aware policies trained on demonstrations on both benchmarks and matching hand-written expert controllers on \bench.
Ablations further show that smaller models make significantly better decisions when given physical quantities rather than raw force values.
\end{abstract}

\section{Introduction}
\label{sec:intro}

Frontier multimodal models can now control robots directly without task-specific training.
For example, GPT-6 Astra runs behind an API with frozen weights and still controls a robot zero-shot from images through a pose interface~\cite{zhang2026astra,robocurve2026astra}.
Much of this progress comes from the \emph{harness} that wraps the frozen model and provides semantic action units~\cite{chen2026showharness}, virtual camera views~\cite{wang2026spatialharness}, agent-written tools and code~\cite{ge2026urai,hu2026coap}, or low-level VLA skills~\cite{zhang2026harnessvla}.
However, existing harnesses mainly help the model see and move.

Beyond seeing and moving, many manipulation tasks in the physical world depend on contact forces that images cannot show.
For example, driving a screw requires pressing down while turning until the thread engages, and sorting identical-looking bottles requires comparing their weight.
Existing work brings force into such tasks mainly through learning.
Specifically, force-aware policies~\cite{yu2025forcevla,li2026forcevla2,he2025foar} and benchmarks~\cite{han2026forcebench} add wrist force as a network input to visuomotor policies~\cite{chi2023dp} and vision-language-action (VLA) models~\cite{pi05}.
However, this approach requires demonstrations for every task and updates model weights, so it cannot be applied to an API-only frozen model.
Meanwhile, existing harnesses for frozen models that expose contact typically pass raw wrench and tactile values~\cite{chu2026openrua,tanaka2026taczero} or ask the model to generate force-control code~\cite{burns2024genchip,tan2026forca}.
Yet these designs still leave the model to parse raw sensor numbers or write low-level control without calibrated physical feedback.
We therefore ask: \emph{can a frozen frontier model understand force and use it to complete manipulation tasks that cannot be done without it?}

Answering this question raises two challenges.
First, contact evolves much faster than an API model can decide, so the system must \emph{close the loop on force beneath the model} through general skills that regulate force on their own under the model's direction.
Second, a frozen model cannot interpret raw force streams or choose actions from raw numbers alone, so the system must \emph{return force in a form the model can reason about}.

Therefore, we present \paper, a harness that lets a frozen frontier model use wrist force without demonstrations or training.
As shown in \cref{fig:teaser}, \paper places three layers between the model and the robot.
At the bottom, a \emph{force sensing and calibration} layer estimates hand force from a wrist sensor or joint torques alone and removes sensor offsets.
In the middle, \emph{self-terminating force primitives} address the first challenge by turning common contact skills such as guarded moves, sliding, and insertion into callable tools.
Once called, each primitive regulates force on its own and returns at a contact event.
At the top, a \emph{physics translation layer} addresses the second challenge by converting each primitive's force trace into physical quantities with uncertainty estimates and status flags, such as payload mass, contact height, and friction.
At each turn, the model calls a primitive, reads the returned physical quantities, and decides what to do next.
Together, these three layers form a general set of building blocks that bridges a frozen frontier model and low-level force control.

We evaluate \paper on a real robot arm without a wrist force sensor and across two simulation benchmarks, as summarized in \cref{fig:teaser}.
On the real robot, we design six contact-rich tasks and find that \paper raises the zero-shot success rate of GPT-6 Astra from 3.3\% without force to 90.0\%.
In simulation, the same model with \paper operates from RGB images on the public ForceBench suite~\cite{han2026forcebench} and improves from 70.0\% to 87.5\%, surpassing the force-aware policies trained on that benchmark.
To further compare frozen models with classical controllers and learned force policies under identical conditions, we build \bench, a seven-task MuJoCo benchmark with hidden physical variables.
On \bench, \paper raises the success rate of GPT-6 Astra from 62.1\% to 96.4\%, matching hand-written expert force controllers and far exceeding Diffusion Policy~\cite{chi2023dp}, $\pi_{0.5}$~\cite{pi05}, and ForceVLA~\cite{yu2025forcevla} trained on about 200 demonstrations per task.

\noindent Our contributions are:
\begin{itemize}
  \item We propose \textbf{\paper}, a three-layer harness of force calibration, self-terminating force primitives, and physics translation that bridges an API-only frozen model and low-level force control without demonstrations or training.
  \item We build \textbf{\bench}, a seven-task MuJoCo benchmark with hidden physical variables that includes expert force controllers and learned force policies evaluated on identical seeds.
  \item Experiments on a real robot and two simulated benchmarks show that \paper lets frozen frontier models make full use of force for contact-rich manipulation, outperforming both training-based methods and the same harness without force.
\end{itemize}

\section{Related work}
\label{sec:related}

\paragraph{Harness for embodied AI.}
GPT-6 Astra showed that a frozen frontier model, given images and a pose interface, can act as a zero-shot robot policy~\cite{zhang2026astra}, and systematic evaluations have mapped what such models can and cannot do as embodied policies~\cite{galbot2026astra,robocurve2026astra}.
Following Code as Policies~\cite{liang2023cap}, in which a language model writes robot programs over control primitives, the harness around the frozen model has become a design variable of its own.
Some harnesses change how the model acts, through semantic action units~\cite{chen2026showharness}, a task-agnostic agent harness~\cite{wang2026auriga}, agent-written tools~\cite{ge2026urai}, or code that runs with no model in the loop~\cite{hu2026coap}.
Others change what it perceives, through calibrated perception tools~\cite{li2026k1} or rendered virtual views~\cite{wang2026spatialharness}.
Others add memory of demonstrations and past interaction~\cite{liu2026roboicl}, orchestrate learned policies as skills~\cite{zhang2026harnessvla,galanti2026pigey}, or let a second agent revise the harness itself~\cite{tian2026physevo,wei2026scale}.
However, few of these harnesses consider force: most have no force channel, and the rest mainly pass raw readings or a force cap to the agent~\cite{fu2026capx,diao2026liberoagent,xie2026roboskill,chu2026openrua}.
\paper is instead built around force: it regulates force beneath the model and reports it to the model as physical quantities.

\paragraph{Force for embodied AI.}
Force has been central to manipulation since hybrid position/force control~\cite{raibert1981hybrid} and impedance control~\cite{hogan1985impedance}, and force-torque measurements can identify a payload~\cite{atkeson1986estimation} or locate a contact~\cite{bicchi1993contact}, but such methods are engineered task by task.
Learned policies instead take force as a network input~\cite{he2025foar,noseworthy2025forge,yu2025forcevla,li2026forcevla2,wang2026lift,gan2026taoforce}, and ForceBench~\cite{han2026forcebench} compares integration designs on $\pi_{0.5}$; all of them require demonstrations or simulated training, and weight updates.
Language models have also used force at inference time by reading sensed properties as sentences~\cite{zhao2023matcha}, generating code with force feedback~\cite{burns2024genchip,monwilliams2025ellmer}, tuning the impedance of a base policy~\cite{zhang2026omnivic,zhang2026compliantvla}, or choosing recovery actions above a trained skill~\cite{rah2026forgeplus}.
Recent studies give GPT-6 Astra tactile or force readings directly~\cite{tanaka2026taczero,yang2026visuotactile,wang2026gpttactile,wang2026tactileeta,zhou2026gpt6}, and FORCA~\cite{tan2026forca} has a coding agent write force-feedback programs and revise them between rollouts.
\paper differs in bringing control and perception of force together with no training: primitives regulate force beneath the model, and the model reads calibrated physical quantities after each one.

\section{Method}
\label{sec:method}

\begin{figure*}[t]
  \centering
  \includegraphics[width=\linewidth]{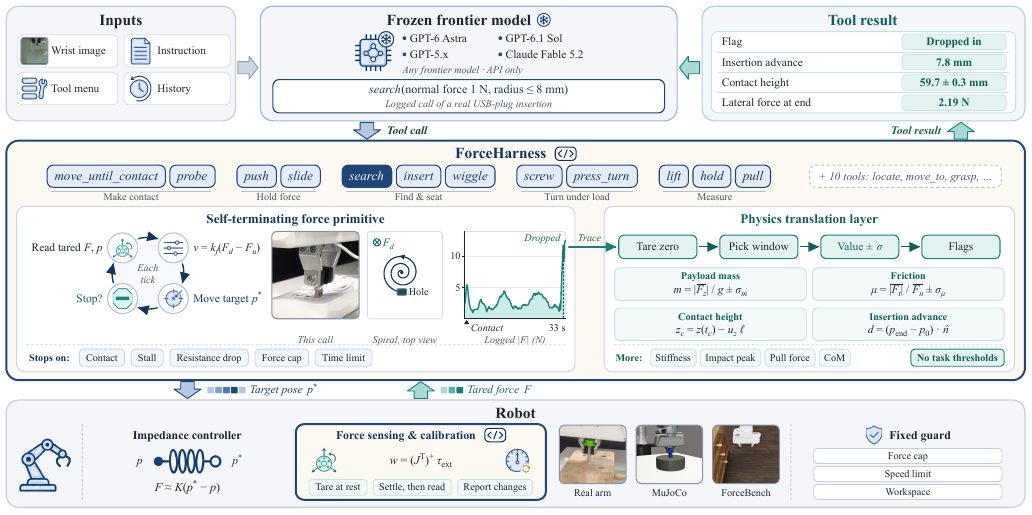}
  \caption{Overview of \paper, illustrated with one logged call.
  Blue marks the command path and teal the force path.
  The frozen model reads its inputs and returns one tool call, here a \tool{search} during a USB-plug insertion.
  The harness runs this call as a self-terminating force primitive that updates the target pose by \cref{eq:spring,eq:admittance} at every tick and stops on a contact event.
  The inset sketches how \tool{search} presses with $F_d$ and follows a spiral until the plug drops into the hole, while the photo and the force trace come from this call.
  Beneath the primitive, the robot's stock impedance controller turns the lead into force, and the bottom layer of \paper estimates and re-zeros the hand force with \cref{eq:wrench}.
  The same harness drives a real arm and two simulators, and a fixed guard bounds what any primitive can do.
  Fixed estimators turn the trace into physical quantities with uncertainties and flags, which return to the model as the tool result.}
  \label{fig:overview}
\end{figure*}

\subsection{Overview}
\label{sec:overview}

\paper sits between a frozen model and a robot and has three layers, shown in \cref{fig:overview}.
The bottom layer, \emph{force sensing and calibration}, turns the robot's raw signals into a calibrated estimate of the force that the environment applies to the hand.
The middle layer is a menu of \emph{self-terminating force primitives}, contact skills that regulate force on their own and stop at a contact event.
The top layer, the \emph{physics translation layer}, converts the force trace of each primitive into physical quantities with uncertainties.

Because the model is reached only through an API and needs seconds for each decision, force enters the loop twice.
Inside a primitive it controls contact at the rate of the controller, and in the tool result it informs the model's next decision.
None of the layers assumes a particular arm, and \suppsec{sec:supp-primitives}{A} of the supplementary material lists the constants of each robot and simulator.

\subsection{Force sensing and calibration}
\label{sec:sensing}

The bottom layer provides one signal to the layers above, the force $F$ that the environment applies to the hand.
A wrist force/torque sensor measures it directly.
Many arms have no such sensor, so the layer can also recover the force from the external joint torques $\tau_{\mathrm{ext}}$ that a torque-controlled arm estimates.
It solves for the wrench $w$ at the flange by least squares, then removes a zero point $b$ and flips the sign,
\begin{equation}
  w = (J^{\top})^{+}\,\tau_{\mathrm{ext}}, \qquad F = -(w_F - b),
  \label{eq:wrench}
\end{equation}
where $J$ is the flange Jacobian and $w_F$ is the force part of $w$.

Such an estimate has little noise, but its zero point drifts as the arm moves and changes pose, and no filter removes a bias that changes this slowly.
The layer therefore controls when it reads rather than how it filters.
It \emph{tares at rest}, re-zeroing $b$ before each empty-hand primitive so that the zero is taken close to the pose of the coming contact.
After each primitive it lets the arm \emph{settle, then reads}, relaxing the controller onto the measured pose and averaging the force over a short still window.
Finally, it \emph{reports changes}, giving each still reading relative to a reference from the same episode, such as the empty hand or the moment of the grasp, so that the model compares readings instead of trusting absolute values.
In simulation, the wrist sensor is instead zeroed at reset and after reorienting the tool.

\begin{table}[t]
  \caption{The twelve force primitives. Rules separate the five groups, which make contact, hold force, find and seat a part, turn under load, and measure. \suppsec{sec:supp-prim-tools}{A.4} of the supplementary material specifies them and the other tools.}
  \label{tab:primitives}
  \centering
  \scriptsize
  \setlength{\tabcolsep}{3pt}
  \renewcommand{\arraystretch}{0.95}
  \begin{tabular}{@{}>{\raggedright\arraybackslash}p{0.30\linewidth}>{\raggedright\arraybackslash}p{0.30\linewidth}>{\raggedright\arraybackslash}p{0.34\linewidth}@{}}
    \toprule
    Tool & What it does & Quantities returned \\
    \midrule
    \tool{move\_until\_contact}, \tool{probe} & Approach until contact & Contact height, impact, stiffness \\
    \midrule
    \tool{push} & Hold a normal force & Mean normal force \\
    \tool{slide} & Slide while pressing & Normal force, friction \\
    \midrule
    \tool{search} & Spiral while pressing until the part drops & Advance, lateral force \\
    \tool{insert} & Push in, slowing as force rises & Advance, lateral force \\
    \tool{wiggle} & Press while rocking & Outcome \\
    \midrule
    \tool{screw} & Press and turn with the thread & Advance per turn, side load \\
    \tool{press\_turn} & Press while turning & Resisting torque, depth \\
    \midrule
    \tool{lift}, \tool{hold} & Hold the load still & Mass, center-of-mass offset \\
    \tool{pull} & Pull along the free path & Pull force, path radius \\
    \bottomrule
  \end{tabular}
\end{table}

\subsection{Self-terminating force primitives}
\label{sec:primitives}

A primitive packages one contact skill as a tool that the model calls with a few parameters, such as the direction to move and the force to hold.
\Cref{tab:primitives} lists the twelve force primitives in five groups, and ten further tools move the hand, grip, perceive, or end the episode.
Three design choices let a model that decides only every few seconds use these tools.

\paragraph{Force through the impedance controller.}
Most arms accept position targets rather than forces, and their impedance controller acts as a spring between the hand and the target.
A primitive therefore applies force $-F$ by letting a target pose $p^{*}$ lead the measured pose $p$,
\begin{equation}
  -F \approx K\,(p^{*} - p), \qquad \lVert p^{*} - p \rVert \le L,
  \label{eq:spring}
\end{equation}
where $K$ is the stiffness of the controller and the clamp $L$ on this lead bounds the force that any primitive can build.
Every primitive thus runs on the stock controller and leaves the low-level control of the robot unchanged.

\paragraph{One shared loop.}
All primitives run in the same loop and differ only in a small control law.
At every tick the law reads the tared force and the pose and returns an increment to $p^{*}$, and the loop clamps the lead, sends the new target, and checks for contact events to decide whether to stop.
To hold a force $F_d$ along a direction $u$, the law turns the force error into a target velocity, which is admittance control,
\begin{equation}
  v = \mathrm{clip}\big(k_f\,(F_d - F_u),\, \pm v_{\max}\big), \quad \Delta p^{*} = u\,v\,\Delta t,
  \label{eq:admittance}
\end{equation}
where $F_u = -F \cdot u$ is the reaction against $u$.
The target keeps moving until the error vanishes, so the outer law integrates the force error while the inner spring supplies the force.

\paragraph{Self-termination.}
A primitive ends on an event rather than on a clock.
It stops as soon as one of the contact events listed in \cref{fig:overview} occurs or its own goal is met, and a force cap and a time limit stop it if nothing else does.
It then unloads the spring and holds the hand still while the model thinks.
This division of work is what makes a slow decision rate workable.
The model decides what to do and how hard to press, the primitive decides when to stop, and a fixed guard beneath both enforces limits that the model cannot change.

\subsection{A representative primitive: search}
\label{sec:search}

We illustrate the design with \tool{search}, the primitive in \cref{fig:overview}, which finds a hole whose position the model knows only roughly.
The model chooses how hard to press, how wide to search, and how deep the part must sink.
The primitive moves toward the surface until it touches it and then keeps pressing with the target force $F_d$ through \cref{eq:admittance}, while the target travels outward on an Archimedean spiral in the plane of the surface, as the inset of \cref{fig:overview} shows.
When the part passes over the hole, the reaction falls, the spiral pauses, and the same law drives the part down.
The primitive reports that the part dropped in once it has sunk past the set depth, or that no hole was found if the spiral leaves the search range first, and in both cases it also reports how far the part advanced and the lateral force at the end.
Most other primitives follow the same pattern of one force law along the contact normal plus their own motion and stop conditions, and \suppsec{sec:supp-prim-tools}{A.4} of the supplementary material specifies them one by one.

\subsection{Physics translation layer}
\label{sec:physics}

A frozen model has not learned to read raw force, and the hundreds of samples that a primitive logs give it little basis for its next decision.
The top layer therefore translates each force trace into a few physical quantities, the numbers a person would read off the trace, and returns them as the tool result.
Its estimators are deliberately simple and fixed, built only from averages and least-squares fits with propagated errors.
Every quantity follows the recipe shown as a strip in \cref{fig:overview}.
The layer removes the zero and any held load, keeps only the window in which the quantity has a single meaning, and reports the value with its uncertainty $\sigma$ and with flags that warn, for instance, of lost contact.
It contains no task thresholds, so it states how heavy a load is or how far a plug advanced but never whether that is enough, which is left to the model.

For example, payload mass comes from a still hold and friction from the part of a slide that stays in contact,
\begin{align}
  m &= \lvert \bar{F}_z \rvert / g, & \sigma_m &= \sqrt{s_z^2/n + \sigma_b^2}\,/\,g, \label{eq:mass}\\
  \mu &= \overline{\lvert F_t \rvert} / \bar{F}_n, & (\sigma_\mu/\mu)^2 &= \big(\sigma_t/\overline{\lvert F_t \rvert}\big)^2 + (\sigma_n/\bar{F}_n)^2, \label{eq:friction}
\end{align}
where $s_z$ is the sample standard deviation of the vertical force $F_z$ over the $n$ samples, $\sigma_b$ is the residual zero bias, and $\sigma_t$ and $\sigma_n$ are the standard errors of the tangential force $F_t$ and the normal force $F_n$.
Contact height is corrected for the lag of the contact detector, stiffness is the least-squares slope of normal force over penetration, and insertion advance carries a flag that says whether the part dropped in or jammed.
\suppsec{sec:supp-phys-quantities}{B.2} of the supplementary material gives every quantity.

\subsection{Model in the loop}
\label{sec:loop}

\Cref{fig:overview} shows one turn of this loop from a real USB-plug insertion.
At every turn the model reads the instruction, the tool menu, the call history, and the latest observations, and it answers with one tool call.
Instructions are written by hand for each task and never reveal the hidden variables of an episode.
A task usually starts without force, as the model points at a target in the image, \tool{locate} turns that pixel into a 3D point from depth, and position moves bring the hand close.
The model then switches to force primitives and sets their parameters from what it has seen and read so far.
Each primitive stops on its own, and its force comes back translated into physical quantities as the tool result.
From these quantities the model judges what happened, for example whether a part has seated or needs another attempt, and it chooses the next call until it ends the episode.
Calibration and safety stay beneath the model, because the bottom layer keeps the force zeroed by the rules of \cref{sec:sensing} and the fixed guard limits force, speed, and workspace for every primitive whatever the model asks.

\section{The \bench benchmark}
\label{sec:bench}

Public force benchmarks such as ForceBench~\cite{han2026forcebench} are built to train and compare learned policies.
We complement them with \bench, a benchmark of seven manipulation tasks in robosuite on MuJoCo.
Its tasks come from real-world contact scenarios and add fine manipulation that ForceBench does not cover.

\paragraph{Tasks.}
Each task hides one physical variable that decides success, as listed in \cref{tab:bench}.
In Weight and CoM the robot must judge a load that it is holding, and in Wipe and Stack it must find a surface whose height is uncertain.
Peg asks it to find a hole by touch, while Latch and Lock ask it to operate a mechanism without damaging it.
The instruction states the range of the hidden variable but never its value, which is drawn from the episode seed, so every method faces the same episodes.

\begin{table}[t]
  \caption{The seven tasks of \bench. Success is checked in code on the final state, and \suppsec{sec:supp-taskcards}{C.2} of the supplementary material gives the full task cards.}
  \label{tab:bench}
  \centering
  \footnotesize
  \setlength{\tabcolsep}{2.5pt}
  \begin{tabular}{@{}lp{0.47\linewidth}p{0.36\linewidth}@{}}
    \toprule
    Task & Hidden variable & What force must do \\
    \midrule
    Weight & Cube mass, 0.15 or 0.60\,kg & Weigh the held cube \\
    CoM & Bar's center of mass, centered or 4\,cm off & Measure the held moment \\
    Wipe & Table height, up to about 6\,cm low & Press 5 to 15\,N on 12 spots \\
    Stack & Lower cube's top, known to $\pm$1\,cm & Feel contact, place gently \\
    Peg & Hole position, 15\,mm error per axis & Search until the peg drops \\
    Latch & Unlock angle, 15 to 35$^\circ$ & Press to release, then pull \\
    Lock & Locked, or stiff and freed at 6--11\,N & Pull below 15\,N \\
    \bottomrule
  \end{tabular}
\end{table}

\paragraph{Setting and references.}
A Panda arm runs operational-space control, and force is the wrist sensor reading with added noise and a per-episode bias.
To isolate force from visual perception, a frozen model reads object poses as noisy text instead of images, while the real robot and ForceBench cover camera input.
\bench ships with three kinds of reference.
The expert force controller is a hand-written state machine for each task, and the position-only controller makes the same motions without using force.
The learned policies are Diffusion Policy~\cite{chi2023dp} and $\pi_{0.5}$~\cite{pi05}, each trained with and without force input, together with a controlled multi-task variant of ForceVLA~\cite{yu2025forcevla}, and all of them learn from about 200 expert demonstrations per task.
Besides success, \bench counts damage, an episode in which the force on a handle passes its damage threshold.
\suppsec{sec:supp-benchmark}{C} of the supplementary material gives the task cards, success checks, and seeds, and \suppsec{sec:supp-baselines}{G} gives the training recipes.

\section{Experiments}
\label{sec:exp}

\begin{figure*}[t]
  \centering
  \includegraphics[width=\linewidth]{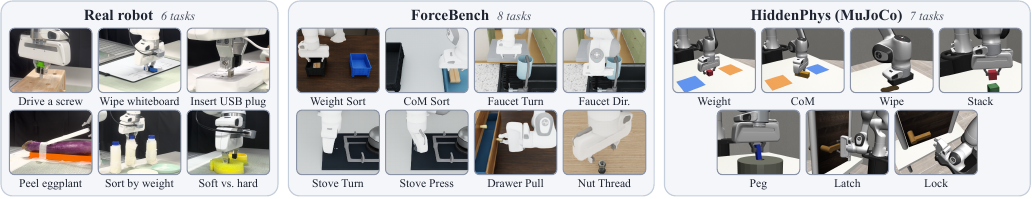}
  \caption{The three experimental settings, with one frame per task, cover six tasks on a real arm without a wrist force sensor, the eight tasks of ForceBench in Isaac Lab, and the seven tasks of \bench in MuJoCo. The colored zones in Weight and CoM are drawn for display only.}
  \label{fig:settings}
\end{figure*}

We evaluate \paper on a real robot, on the public ForceBench suite, and on \bench, shown in \cref{fig:settings}.
In each setting, the same frozen model runs with \paper and with a \emph{no-force control}, which keeps the harness and its safety limits but removes force primitives and force readings.

\subsection{Real-robot experiments}
\label{sec:real}

\paragraph{Experimental setting.}
The robot is a Franka FR3 with a parallel gripper and no wrist force/torque sensor, so force is estimated from joint torques with \cref{eq:wrench}.
Its stock joint-impedance controller runs unchanged, and a wrist RGB-D camera and an exterior RGB camera give the observations.
We test GPT-6 Astra in both conditions and check transfer to GPT-6.1 Sol with \paper, both through the Codex CLI at medium reasoning effort.
Each model receives the text history and the images of the last two turns.

\begin{table}[t]
  \caption{Real-robot results: successes out of 10 episodes per condition, with the same instructions. For Astra, every task differs at $p<0.01$ by Fisher's exact test.}
  \label{tab:real}
  \centering
  \footnotesize
  \setlength{\tabcolsep}{3pt}
  \begin{tabular}{@{}lccc@{}}
    \toprule
    & \multicolumn{2}{c}{GPT-6 Astra} & GPT-6.1 Sol \\
    \cmidrule(lr){2-3}\cmidrule(l){4-4}
    Task & No force & \paper & \paper \\
    \midrule
    Wipe a whiteboard & 1/10 & 8/10 & 6/10 \\
    Drive a screw & 0/10 & 9/10 & 7/10 \\
    Peel an eggplant & 1/10 & 9/10 & 9/10 \\
    Insert a USB plug & 0/10 & 8/10 & 7/10 \\
    Sort three bottles by weight & 0/10 & 10/10 & 8/10 \\
    Tell soft from hard & 0/10 & 10/10 & 10/10 \\
    \midrule
    Total & 2/60 & \textbf{54/60} & 47/60 \\
    \bottomrule
  \end{tabular}
\end{table}

We design six real-world tasks listed in \cref{tab:real}.
Four contact tasks require regulating force during motion, either to wipe all marker writing off a whiteboard and peel a strip of skin from an eggplant, or to seat a screw in a threaded hole and fully insert a USB plug.
The other two tasks test properties that cameras cannot see.
Bottle sorting requires placing the heaviest of three look-alike bottles on a white plate and the lightest on a green plate while leaving the middle one in place, and telling soft from hard requires probing two identically wrapped objects to name each one correctly.
Across these tasks, \paper offers 20 to 21 tools, whereas the no-force control keeps only the ten other tools and \tool{hold} with kinematics-only results, plus a guarded move in peeling that stops at a fixed force without reporting it.
Instructions include procedural hints except in telling soft from hard, where no method is given.
Both conditions share the same safety limits with an operator at the emergency stop.
Each task runs 10 episodes per condition with the same model and instruction after a scene reset, and the operator labels success from the final state.
An external web monitor replays every episode, as \suppsec{sec:supp-monitor}{E} of the supplementary material shows.

\paragraph{Results.}
In \cref{tab:real}, \paper raises GPT-6 Astra from 2/60 without force to 54/60 across all six tasks, while GPT-6.1 Sol reaches 47/60 under the same prompts.
On bottle sorting and telling soft from hard, RGB-D images give no cue to mass or stiffness, leaving the no-force control at 0/10 on both.
On the sensorless FR3, joint-torque estimates drift with arm pose and misstate absolute grams, yet taring at rest, settling before reading, and comparing within-episode changes keep relative order reliable.
This lets Astra score 10/10 on both tasks and Sol score 8/10 and 10/10, even when no probing method is given for soft versus hard.

On the four contact tasks, wrist depth localizes targets only to about 1 to 2\,cm, so position-only steps either miss the surface or hit the collision guard without knowing the contact load, succeeding in only 2 of 40 episodes.
With \paper, force acts both inside and between primitives, as \cref{fig:teaser} illustrates.
Inside each call, guarded moves and admittance control absorb height and pose error, holding steady contact across the flat board or curved eggplant and spiraling at 1\,N until a resistance drop guides the USB plug into its port.
Between calls, tool results and images guide adaptation: remaining ink after an 8\,N wipe prompts Astra to retry at 10\,N, rising side load during screwing triggers an in-place release before binding, and a 2.77\,N mean force over a 51\,mm peel tells Astra to keep its 6\,N target.
Calibrated joint-torque sensing, self-terminating primitives, and physical quantities thus make zero-shot contact manipulation reliable without a wrist sensor.

\begin{table*}[t]
  \caption{ForceBench success in percent. The trained rows are selected from the ForceBench paper and not reproduced, with 200 demonstrations and 200 evaluation episodes per task, and \suppsec{sec:supp-forcebench}{H} of the supplementary material lists all ten. The zero-shot rows are ours, with 10 episodes per task.}
  \label{tab:forcebench}
  \centering
  \footnotesize
  \setlength{\tabcolsep}{4.5pt}
  \begin{tabular}{@{}ll cccccccc c@{}}
    \toprule
    & Method & Weight & CoM & Faucet & Faucet dir. & Stove & Stove press & Drawer & Nut & Mean \\
    \midrule
    \multirow{6}{*}{Trained} & $\pi_{0.5}$, no force & 54.0 & 45.5 & 9.5 & 31.5 & 8.5 & 1.0 & 4.5 & 31.5 & 23.25 \\
     & $\pi_{0.5}$ + Prompt Fusion & 88.0 & 56.5 & 21.0 & 41.5 & 16.0 & 2.5 & 23.0 & 34.0 & 35.31 \\
     & $\pi_{0.5}$ + Dual-Path Fusion & 99.5 & 72.0 & 22.0 & 30.0 & 17.5 & 19.0 & 4.5 & 16.5 & 35.13 \\
     & $\pi_{0}$~\cite{black2024pi0} & 61.5 & 39.0 & 3.5 & 27.0 & 7.0 & 5.0 & 7.5 & 36.0 & 23.31 \\
     & ForceVLA & 35.5 & 45.5 & 12.5 & 14.0 & 10.0 & 12.0 & 20.0 & 26.0 & 21.94 \\
     & ForceFlow~\cite{zhang2026forceflow} & 51.5 & 3.0 & 0.0 & 0.5 & 0.0 & 0.0 & 0.0 & 0.0 & 6.88 \\
    \midrule
    \multirow{5}{*}{Zero-shot} & GPT-6 Astra, no force & 80 & 50 & 100 & 50 & 100 & 80 & 0 & 100 & 70.00 \\
     & GPT-6 Astra + \paper & 100 & 70 & 100 & 50 & 100 & 90 & 100 & 90 & \textbf{87.50} \\
     & GPT-6.1 Sol, no force & 90 & 50 & 90 & 30 & 100 & 60 & 0 & 100 & 65.00 \\
     & GPT-6.1 Sol + \paper & 100 & 60 & 100 & 40 & 100 & 100 & 100 & 100 & \textbf{87.50} \\
     & Claude Fable 5.2 + \paper & 100 & 80 & 100 & 20 & 100 & 50 & 100 & 50 & 75.00 \\
    \bottomrule
  \end{tabular}
\end{table*}

\subsection{ForceBench}
\label{sec:forcebench}

\paragraph{Experimental setting.}
ForceBench~\cite{han2026forcebench} is a public Isaac Lab suite of eight contact-rich tasks, where we keep its evaluator and run its first ten seeds.
Our harness serves as the policy with the same primitives, estimators, and a few suite-specific tools, reading two RGB views, object sizes, and key heights without object coordinates.
Instructions state task facts that demonstrations expose, such as the two mass classes, and the no-force control adds a height-following turn tool so it lacks force rather than motion.
GPT-6 Astra and GPT-6.1 Sol run at low reasoning effort, and Claude Fable 5.2 runs with \paper at high effort.

\paragraph{Results.}
In \cref{tab:forcebench}, Astra and Sol reach 87.5\% with \paper and Fable reaches 75.0\%, above 35.31\% for the best trained policy.
Since Astra and Sol already score 70.0\% and 65.0\% without force through the model, harness, and instructions, that gap is not from force alone.
Within our harness, force brings paired gains at McNemar's $p$ of 0.009 and 0.001 mainly on Drawer Pull from 0/10 to 10/10, while other tasks do not separate at 10 episodes, showing clean transfer when subtle resistance decides success.

Two protocol notes apply.
First, tools and perception thresholds were tuned on eight of the ten seeds, and Stove Press and Nut Thread were revised after initial runs.
Second, on Faucet Turn, Stove Turn, and Drawer Pull the public evaluator omits checks such as release and clearance.

\subsection{\bench}
\label{sec:mujoco}

\paragraph{Experimental setting.}
GPT-6 Astra runs at low reasoning effort and Claude Fable 5.2 runs at high effort, each with at most 40 turns, 80 calls, and 300 simulated seconds per episode.
All rows of \cref{tab:main} share the 20 test seeds per task, the hidden variables, and the success checks of \cref{sec:bench}.
Only the learned policies also see two camera views, which show them the true hole on Peg.
Because learned policies have no force cap, we report damage for every row.
The classical rows confirm that the tasks depend on force: over 50 seeds in \suppsec{sec:supp-forcedep}{C.4}, the position-only controller falls from 49 expert successes to none on Wipe and from 50 to 25 on Weight.

\begin{table*}[t]
  \caption{Results on \bench as successes out of 20 test seeds per task. Damage counts episodes in which the force on a handle passes its damage threshold. Demos is the number of demonstrations per task, and Code marks task-specific code. DP trains one model per task, while $\pi_{0.5}$ and ForceVLA train one multi-task model, each with one training seed. Shift gives successes out of 80 on four tasks, in distribution and under shift.}
  \label{tab:main}
  \centering
  \footnotesize
  \setlength{\tabcolsep}{3pt}
  \begin{tabular}{@{}ll ccccccc c c c c c@{}}
    \toprule
    & Method & Weight & CoM & Wipe & Stack & Peg & Latch & Lock & Total & Damage & Demos & Code & Shift \\
    \midrule
    \multirow{2}{*}{Classical} & Expert force controller & 20 & 19 & 20 & 20 & 18 & 19 & 20 & 136 & 1 & 0 & \checkmark & 77 $\rightarrow$ 68 \\
     & Position only & 10 & 10 & 0 & 17 & 5 & 13 & 16 & 71 & 6 & 0 & \checkmark & 25 $\rightarrow$ 22 \\
    \midrule
    \multirow{7}{*}{Learned} & DP & 16 & 17 & 4 & 11 & 15 & 9 & 11 & 83 & 20 & 200 & -- & 52 $\rightarrow$ 40 \\
     & DP + force & 18 & 17 & 7 & 6 & 15 & 1 & 10 & 74 & 13 & 200 & -- & 57 $\rightarrow$ 43 \\
     & DP, images only & 9 & 17 & 4 & 12 & 14 & 0 & 9 & 65 & 25 & 200 & -- & 44 $\rightarrow$ 42 \\
     & DP, images only + force & 16 & 17 & 5 & 10 & 14 & 3 & 9 & 74 & 16 & 200 & -- & 52 $\rightarrow$ 43 \\
     & $\pi_{0.5}$ & 9 & 9 & 14 & 12 & 12 & 2 & 11 & 69 & 24 & 200 & -- & 44 $\rightarrow$ 35 \\
     & $\pi_{0.5}$ + force, Prompt Fusion & 15 & 10 & 7 & 14 & 10 & 0 & 11 & 67 & 18 & 200 & -- & 42 $\rightarrow$ 35 \\
     & ForceVLA & 11 & 10 & 12 & 12 & 15 & 2 & 8 & 70 & 29 & 200 & -- & 48 $\rightarrow$ 36 \\
    \midrule
    \multirow{4}{*}{Ours} & GPT-6 Astra, no force & 17 & 10 & 16 & 20 & 1 & 13 & 10 & 87 & 17 & 0 & -- & 44 $\rightarrow$ 37 \\
     & \textbf{GPT-6 Astra + \paper} & 20 & 20 & 20 & 20 & 15 & 20 & 20 & \textbf{135} & \textbf{0} & 0 & -- & \textbf{75 $\rightarrow$ 69} \\
     & Claude Fable 5.2, no force & 10 & 18 & 17 & 20 & 3 & 18 & 10 & 96 & 11 & 0 & -- & 48 $\rightarrow$ 39 \\
     & \textbf{Claude Fable 5.2 + \paper} & 20 & 20 & 20 & 20 & 15 & 20 & 20 & \textbf{135} & \textbf{0} & 0 & -- & \textbf{75 $\rightarrow$ 71} \\
    \bottomrule
  \end{tabular}
\end{table*}

\begin{table}[t]
  \caption{Form of force. L2, our default, adds $\pm\,\sigma$, window, and flags to L1. Success is the share of episodes passing the task check, decision the share of correct verdicts on Weight, CoM, and Lock. Sol runs 70 episodes per row, the five smaller models 125.}
  \label{tab:repr}
  \centering
  \footnotesize
  \setlength{\tabcolsep}{5pt}
  \begin{tabular}{@{}lcccc@{}}
    \toprule
    & \multicolumn{2}{c}{GPT-6.1 Sol} & \multicolumn{2}{c}{Smaller models} \\
    \cmidrule(lr){2-3}\cmidrule(l){4-5}
    Form & Success & Decision & Success & Decision \\
    \midrule
    No force & 0.63 & 0.73 & 0.18 & 0.57 \\
    Event words & 0.77 & 0.70 & 0.54 & 0.64 \\
    Force plot & 0.90 & \textbf{1.00} & 0.56 & 0.71 \\
    Raw numbers & 0.87 & \textbf{1.00} & 0.63 & 0.69 \\
    Derived signals & 0.87 & 0.97 & 0.69 & \textbf{0.85} \\
    Quantities, L1 & \textbf{0.91} & \textbf{1.00} & 0.64 & 0.84 \\
    + uncertainty, L2 & 0.90 & \textbf{1.00} & \textbf{0.70} & \textbf{0.85} \\
    \bottomrule
  \end{tabular}
\end{table}

\paragraph{Main results.}
In \cref{tab:main}, both Astra and Fable with \paper reach 135/140 with zero handle damage, matching the hand-written expert at 136/140.
All trained policies score only 65/140 to 83/140 with frequent damage, and adding wrist force to Diffusion Policy, $\pi_{0.5}$, or ForceVLA brings no consistent gain because passive conditioning neither teaches active probing nor caps force.

Within our harness, force lifts Astra from 87/140 to 135/140 and Fable from 96/140 to 135/140, with consistent gains across all eight OpenAI models in \suppsec{sec:supp-models}{D.4} of the supplementary material.
Without force, Astra partially copes on Wipe and Weight by small position steps, while Fable takes six times as many calls and Astra overloads half of its Lock handles.
\paper eliminates handle damage and cuts Astra's cost per episode by 45\%, matching task-specific force controllers safely.

\paragraph{Distribution shift.}
We next shift the hidden ranges on four tasks without retraining while updating instructions only for the frozen models, as detailed in \suppsec{sec:supp-shift}{D.7} of the supplementary material.
With \paper, Astra and Fable show relative drops of only 8\% and 5\% to 69/80 and 71/80, matching the fixed-threshold expert at 68/80.
Without force they lose 16\% to 19\% and learned policies lose 17\% to 25\%, aside from image-only Diffusion Policy losing 5\% from a low 44/80 start as its wrist camera sees the Peg hole.

\subsection{Ablation studies}
\label{sec:ablation}

\begin{figure}[t]
  \centering
  \includegraphics[width=\linewidth]{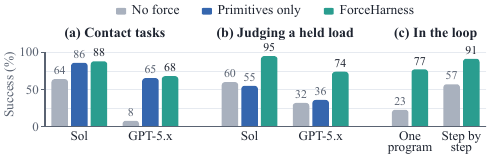}
  \caption{Ablations on \bench. (a) Contact tasks, without Peg and Latch for GPT-5.x. (b) Weight and CoM. (c) GPT-6 Astra, five seeds per task.}
  \label{fig:ablation}
\end{figure}

To see where the gains of \paper come from, we ablate its force primitives, feedback form, and decision loop on \bench.
\Cref{tab:repr} compares seven feedback forms on identical primitives and seeds, \cref{fig:ablation} splits the gains by task group and loop structure, and \suppsec{sec:supp-results}{D} of the supplementary material gives per-task results.

\paragraph{Force primitives.}
The event-words row in \cref{tab:repr} keeps our force primitives but reports only events with $|F|$.
Even without force traces, these primitives raise success from 0.63 to 0.77 for Sol and from 0.18 to 0.54 for smaller GPT-5.x models, mainly on the contact tasks in \cref{fig:ablation}a, while cutting Lock handle damage for smaller models from 22 of 25 episodes to 3.
Self-terminating primitives alone thus secure safe contact execution beneath the model.

\paragraph{Physics translation layer.}
Judging a held load in \cref{fig:ablation}b gives no stop event, so event words fail and force values are needed.
\Cref{tab:repr} compares returning force plots, tared raw numbers, or harness-computed signals and quantities.
Sol reaches 0.87 to 0.91 across all three.
For smaller models, force plots reach only 0.56 success, and raw numbers lift decision accuracy insignificantly from 0.57 to 0.69.
L2 quantities raise decision accuracy to 0.85 at McNemar's $p$ of 0.023 over raw numbers and lift success to 0.70, beating plots at $p$ of 0.005 and gaining modestly over 0.63 for raw numbers since primitives drive contact tasks.
Derived signals, L1, and L2 match closely, showing that smaller models need harness-computed quantities rather than plots or raw samples.

\paragraph{Model in the loop.}
In \cref{fig:ablation}c, we compare turn-by-turn calls with one-shot Code-as-Policies~\cite{liang2023cap} programs written by Astra.
Our force interface lifts one-shot programs from 23\% to 77\% though only our harness was tuned, and turn-by-turn feedback reaches 91\% by adapting after failures on Stack and Peg.

\section{Conclusion}
\label{sec:conclusion}

We presented \paper, a three-layer harness that lets an API-only frozen frontier model use wrist force with no demonstrations or weight updates.
Across a real arm without a wrist sensor, ForceBench, and \bench, keeping force calibration, admittance control, and event termination beneath the model while returning calibrated physical quantities lets frozen models match hand-written expert controllers and surpass policies trained on demonstrations.
Ablations show that self-terminating primitives keep contact safe without force values, physical quantities help smaller models judge a held load, and calling the model after each primitive lets it recover from failures.

\paragraph{Limitations.}
Model queries add seconds between primitives and preclude reactive coordination, a fixed menu limits new motion laws, and our ablations run only in simulation.

{
    \small
    \bibliographystyle{ieeenat_fullname}
    \bibliography{main}
}

\clearpage
\twocolumn[{\centering\Large\textbf{\thetitle}\\\vspace{0.5em}Supplementary Material\\\vspace{1.0em}}]
\appendix
\input{sec/suppl/0_setup}

Real-robot traces in this supplement illustrate the system, and the evaluation is Table~3 of the main paper.

{\small
\tableofcontents
}
\addtocontents{toc}{\protect\setcounter{tocdepth}{1}}

\section{Self-terminating force primitives}
\label{sec:supp-primitives}

This section specifies the force primitives of Sec.~3.3 and Table~1 of the main paper and every other tool of the menu.
All of them share one control loop, so we describe the loop first, then how a position-controlled arm produces force inside it, then the constants that differ between backends, and finally each primitive with a sketch, pseudocode, and one logged call.
The backends are MuJoCo, the robosuite Panda of \bench, ForceBench, the Isaac Lab arm behind that benchmark's evaluator, and the real arm, the FR3 of Sec.~5.1 of the main paper.
The constants are taken from the code of each backend.

Throughout, $p$ is the measured tool center point and $p^{*}$ the target, so $p^{*}-p$ is the lead.
$F$ is the force that the environment applies to the hand, in the base frame, with the zero and the held-load tare $b_c$ removed, and $F_{\mathrm{raw}}$ is the same force before $b_c$ is removed.
A unit normal $n$ points into the surface, so that $F_n=-F\cdot n$ is positive while the hand presses.
The slow and fast target speeds are $v_{\mathrm{near}}$ and $v_{\mathrm{free}}$, and one tick lasts $\Delta t$.
In the pseudocode, teal marks force terms and navy marks targets and motion, as in Figure~2 of the main paper.

\subsection{The shared loop}
\label{sec:supp-prim-loop}

\begin{figure*}[!t]
  \centering
  \begin{minipage}[t]{0.435\textwidth}\vspace{0pt}
    \includegraphics[width=\linewidth]{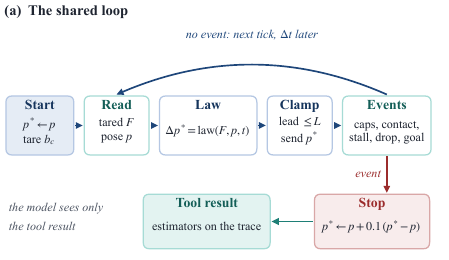}
  \end{minipage}\hfill
  \begin{minipage}[t]{0.535\textwidth}\vspace{0pt}
  \begin{pcode}{(b)\enspace\Fn{Run}, the loop of every moving tool}
  \PL{0}{\Kw{input} control law, time limit $T_{\max}$, lead limit $L$, stops $S$ that the tool asks for}
  \PL{0}{$\Tq{p^{*}}\leftarrow p$, reset the event detector \Cm{no lead carries over}}
  \PL{0}{\Kw{if} the gripper is closed \Kw{and} the still reading is load-like \Kw{then} $\Fq{b_c}\leftarrow$ that reading}
  \PL{0}{\Kw{for} $k=1,\dots,T_{\max}/\Delta t$ \Kw{do}}
  \PL{1}{$(\Tq{\Delta p^{*}},\Delta\psi,\mathit{done})\leftarrow\mathrm{law}(\Fq{F},p,t)$}
  \PL{1}{$\Tq{p^{*}}\leftarrow p+\min\!\big(1,\,L/\lVert\Tq{p^{*}}+\Tq{\Delta p^{*}}-p\rVert\big)\,(\Tq{p^{*}}+\Tq{\Delta p^{*}}-p)$ \Cm{clamp the lead}}
  \PL{1}{send $\Tq{p^{*}}$ and the yaw target to the impedance controller, wait one tick}
  \PL{1}{$\Fq{F}\leftarrow\Fq{F_{\mathrm{raw}}}-\Fq{b_c}$, \ $E\leftarrow\Fn{Detect}(\Fq{F},\tau,p,\Tq{\Delta p^{*}})$ \Cm{\cref{tab:supp-prim-events}}}
  \PL{1}{\Kw{if} \code{force\_cap} or \code{torque\_cap} $\in E$, \Kw{or} $E\cap S\neq\emptyset$ \Kw{then stop} with that label}
  \PL{1}{\Kw{if} $\mathit{done}$ \Kw{then stop} with \code{completed}}
  \PL{0}{\Kw{if} the ticks ran out \Kw{then} label \code{timeout}}
  \PL{0}{\Kw{if} stopped by an event \Kw{then} $\Tq{p^{*}}\leftarrow p+0.1\,(\Tq{p^{*}}-p)$ \Cm{unload the spring}}
  \PL{0}{hold still, keep at most 160 rows, run the estimators \Cm{the tool result}}
  \end{pcode}
  \end{minipage}
  \caption{The loop behind every primitive. (a)~Each tick reads the force and pose, applies the law of the tool, clamps the lead, and checks for events. Only an event or the law's own goal ends the call, and only the resulting tool result reaches the model. (b)~Pseudocode. Each card of \cref{sec:supp-prim-tools} gives the law and the stops of one tool.}
  \label{fig:supp-prim-loop}
\end{figure*}

Every tool that moves the robot is a small control law plus one call of the loop in \cref{fig:supp-prim-loop}.
The loop starts from $p^{*}=p$, so no lead carries over from the previous call, and with a closed gripper it first re-estimates the held-load tare from a still, load-like reading.
At every tick the law reads the tared force and the pose and returns an increment of the target.
The loop clamps the lead to the limit $L$ of the tool, which bounds the force that any primitive can build, sends the target to the impedance controller, and runs the event detector.
Force and torque caps stop every call, while the other events stop a call only if the tool asked for them.
After a stop on an event, the target returns to within a tenth of its lead, which unloads the spring, and the hand then holds still while the model thinks.
The loop keeps at most 160 evenly spaced rows of the trace for the estimators of \cref{sec:supp-physics}, and the peak force is tracked before thinning.

A reading $f$ counts as load-like when $f_z<-1$\,N, $\lVert f\rVert<15$\,N, and $\lVert f_{xy}\rVert\le\max(1\,\mathrm{N},\,0.5\,|f_z|)$.
The tare $b_c$ is cleared by \tool{grasp} and \tool{release}, set by \tool{lift} from its still window, and replaced at the end of any still, completed call that holds a load-like reading.
On the real arm, the runner also reads the force still for a nominal 0.5\,s after each call, which gives the still reading of Sec.~3.2 of the main paper.

\subsection{Force from lead, and contact events}
\label{sec:supp-prim-lead}

No backend offers a force command at this level.
Each accepts a position target and behaves as a spring between hand and target.
MuJoCo runs an operational-space law with pose-dependent stiffness, ForceBench uses inverse kinematics with fixed joint gains, and the real arm runs its stock joint-impedance controller with fixed gains, so all compliance comes from the outer loop and the lead.
With the hand blocked along a direction $u$,
\begin{equation}
  F_u \approx K\,(p^{*}_u-p_u), \qquad \lVert p^{*}-p\rVert\le L,
  \label{eq:supp-spring}
\end{equation}
so the lead limit bounds the force at $KL$.
On the real arm the measured $K$ is 0.75 to 1\,N/mm, because pressing with a 15\,mm lead read 10 to 11\,N and free-air steps gave about 1\,N/mm, so 8\,mm of lead holds at most about 8\,N and 25\,mm about 18\,N.

To hold $F_d$ along $u$, the primitives run admittance control on top of this spring,
\begin{equation}
  p^{*}_{k+1}=p^{*}_{k}+u\,\mathrm{clip}\big(k_f(F_d-F_{u,k}),\,\pm v_{\max}\big)\,\Delta t,
  \label{eq:supp-admit}
\end{equation}
with $F_u=-F\cdot u$.
The target integrates the force error, so the steady error is zero while the lead stays below $L$.
Unsaturated and blocked, the error shrinks by a factor $1-\lambda$ per tick with $\lambda=Kk_f\Delta t$.
Computed from the constants rather than measured, $\lambda$ for \tool{push} is 0.17 to 0.23 on the real arm, 0.09 to 0.14 in MuJoCo, and 0.15 to 0.37 in ForceBench, so the force approaches its target without oscillation.
On the real arm each tick therefore corrects about a fifth of the error and moves the target at most 0.2\,mm, and the force rises at 3 to 4\,N/s.

Two real-arm primitives, \tool{screw} and \tool{wiggle}, press with a fixed lead instead of a force loop.
They keep the target a fixed lead $\ell=F_d/K_z$ below the hand, with $K_z=750$\,N/m.
Both turn the wrist, which shifts the zero of the force estimate, and a fixed lead presses with about $K_z\ell$ independently of that shift.

The events that stop a primitive come from fixed thresholds with hysteresis on $\tilde F$, the mean of the last $N_f$ values of $\lVert F\rVert$, and nothing is learned.
\Cref{tab:supp-prim-events} gives the rules.
Contact is suppressed in \tool{press\_turn}, and with a closed gripper in \tool{lift}, \tool{hold}, \tool{pull}, and \tool{turn}, because a held load or handle is not a new contact.
On the real arm, the position moves, the turns, \tool{lift}, and \tool{hold} are guarded against collisions, so the detector there uses $f_c=8$\,N and the runner also watches $\lVert F_{\mathrm{raw}}\rVert$, which a held load cannot hide.

\begin{table}[t]
  \caption{Contact events. Thresholds act on the filtered force $\tilde F$ except for the caps, which act on every raw tick. \Cref{tab:supp-prim-constants} gives $f_c$, $f_r$, $m$, and the caps of each backend.}
  \label{tab:supp-prim-events}
  \centering
  \footnotesize
  \setlength{\tabcolsep}{3pt}
  \begin{tabular}{@{}p{0.31\linewidth}p{0.44\linewidth}p{0.19\linewidth}@{}}
    \toprule
    Event & Rule & Stops \\
    \midrule
    \code{contact\_start} & $\tilde F\ge f_c$ on $m$ ticks & if the tool asks \\
    \code{contact\_end} & $\tilde F<f_r$ on $m$ ticks & reported \\
    \code{force\_cap} & $\lVert F\rVert\ge f_{\mathrm{cap}}$ & always \\
    \code{torque\_cap} & $\lVert\tau\rVert\ge t_{\mathrm{cap}}$ & always \\
    \code{stall} & in contact, the target moved 2\,mm in six ticks and the hand under 0.35 of that & reported, tools test stalls their own way \\
    \code{slip} & gripper closed, $\tilde F$ fell below half its six-tick peak, which exceeded $f_c$ & reported \\
    \code{resistance\_drop} & in contact, $\tilde F$ fell below 0.65 of a peak of at least 4\,N while the hand moves & if the tool asks \\
    \bottomrule
  \end{tabular}
\end{table}

\subsection{Constants of each backend}
\label{sec:supp-prim-backends}

The design of the primitives is the same on every backend, while speeds, gains, and limits follow from each robot's controller and sensing.
\Cref{tab:supp-prim-constants} collects these constants, and each card in \cref{sec:supp-prim-tools} gives the lead limit of its tool.
The real arm moves its target five to twelve times more slowly than simulation and allows 8\,mm of lead in free motion instead of 30\,mm.

\begin{table*}[t]
  \caption{Constants of each backend. All real-arm speeds and ticks are nominal and its forces use the torque-based estimate. Its lead limits are those of the workstation, and the real-time computer clamps again.}
  \label{tab:supp-prim-constants}
  \centering
  \footnotesize
  \setlength{\tabcolsep}{4pt}
  \begin{tabular}{@{}p{0.25\linewidth}p{0.225\linewidth}p{0.2\linewidth}p{0.27\linewidth}@{}}
    \toprule
    Constant & MuJoCo & ForceBench & Real arm \\
    \midrule
    Tick $\Delta t$ & 0.05\,s & 1/15\,s & 0.05\,s nominal, about 66\,ms measured \\
    $v_{\mathrm{near}}$, $v_{\mathrm{free}}$ & 20, 100\,mm/s & 20, 100\,mm/s & 4, 8\,mm/s \\
    $k_f$: \tool{push}, \tool{pull}, \tool{press\_turn} / \tool{slide}, \tool{search} & 0.0045 / 0.0015\,(m/s)/N & same & same, no \tool{press\_turn} \\
    Stiffness command & $k_p$ 150, 40 on a soft axis & joint gains 80 and 12, fixed & 750\,N/m and 15\,N$\cdot$m/rad, fixed \\
    Effective $K$ when blocked & about 3\,N/mm at $k_p$ 150, 0.4 to 0.6 at 40 & 1.23\,N/mm near, about 0.5 far & 0.75 to 1\,N/mm \\
    Lead limit $L$: free, guarded & 30, 12\,mm & 30, 12\,mm & 8, 8\,mm \\
    Lead limit $L$ while pressing & by tool, in each card & same & by tool, at most 25\,mm once the raw force changed by 3\,N \\
    Filter $N_f$, samples $m$ & 4 ticks, 1 & 4 ticks, 1 & 1 tick, 3 after a 0.3\,s grace \\
    $f_c$, $f_r$ & 3\,N, 1\,N & 2\,N, 1\,N & 5\,N or 8\,N if guarded, $0.5f_c$ \\
    Caps $f_{\mathrm{cap}}$, $t_{\mathrm{cap}}$ & 25\,N tared, 3\,N$\cdot$m & same & 25\,N, raw on the real-time computer, 3\,N$\cdot$m \\
    Force target ceiling & 20\,N & 20\,N & 20\,N, lower per tool \\
    Rates of \tool{rotate\_z}, \tool{turn}, \tool{point\_tool} & 30, 10, 30$^{\circ}$/s & same & 8, 2, 2$^{\circ}$/s \\
    Position tolerance & 3\,mm, 0.1\,mm on Lock & 3\,mm & 3\,mm reported, 1\,mm internal \\
    Added force noise & 0.5\,N per tick, 0.3\,N per episode & none & none, about 0.03\,N static \\
    \bottomrule
  \end{tabular}
\end{table*}

\subsection{The primitives one by one}
\label{sec:supp-prim-tools}

The cards below follow the five groups of Table~1 of the main paper and then the tools that turn the hand or move it by position.
Each shows a sketch of the mechanism, the control law and stops as pseudocode that runs inside \Fn{Run}, and one logged call whose tool-result excerpt is quoted as printed.
Every moving tool also stops on a cap or a timeout.
On the real arm the runner adds two stops, \code{path\_deviation} when the hand leaves its path by more than 15\,mm, or 20\,mm for \tool{turn} and 10\,mm for \tool{screw}, and \code{stalled} when the hand gains under 0.5\,mm in 3.5\,s against more than 2\,mm of lead, and a \code{force\_cap} or a deviation above 30\,mm ends the episode.

\begin{figure*}[!t]
  \centering
  \begin{minipage}[t]{\cardgfx}\vspace{0pt}\includegraphics[width=\linewidth]{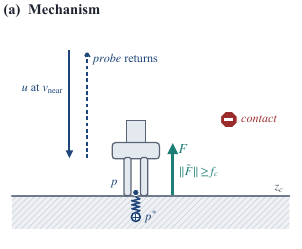}\end{minipage}\hfill
  \begin{minipage}[t]{\cardcode}\vspace{0pt}
  \begin{pcode}{(b)\enspace Pseudocode}
  \PL{0}{\Kw{input} direction $u$, \code{max\_dist}, threshold $f_c$ \Cm{$L=6$\,mm, 15\,mm in ForceBench}}
  \PL{0}{$p_0\leftarrow p$, \Fn{Run} with stops $S=\{\code{contact}\}$ and, every tick,}
  \PL{1}{\Kw{if} $(p-p_0)\cdot u\ge\code{max\_dist}$ \Kw{then} $\mathit{done}$}
  \PL{1}{$\Tq{\Delta p^{*}}\leftarrow\Tq{u}\,v_{\mathrm{near}}\Delta t$ \Cm{the law reads no force}}
  \PL{0}{\Kw{if} $\mathit{done}$ \Kw{then} label \code{no\_contact\_within\_max\_dist}}
  \PL{0}{\Kw{return} contact height, impact peak, stiffness}
  \PL{0}{\Fn{Probe}$(u,\code{amplitude})$: run the above with $\code{max\_dist}=\code{amplitude}$, then \tool{move\_to}$(p_0)$}
  \end{pcode}
  \end{minipage}\hfill
  \begin{minipage}[t]{\cardgfx}\vspace{0pt}\includegraphics[width=\linewidth]{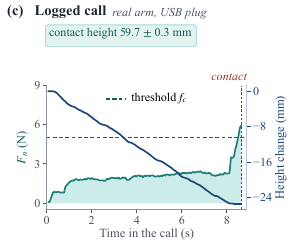}\end{minipage}
  \caption{\tool{move\_until\_contact} and \tool{probe}. (a)~The target leads the hand toward the surface until the filtered force passes $f_c$. (b)~Pseudocode. (c)~The approach before the \tool{search} of Figure~2 of the main paper on the real arm. The estimate reads about 2\,N of drift while the arm moves, and the call stops when $F_n$ passes $f_c$.}
  \label{fig:supp-prim-contact}
\end{figure*}

\paragraph{Making contact.}
\tool{move\_until\_contact} walks the target along $u$ at $v_{\mathrm{near}}$ until the detector reports contact, and its law never reads force, as \cref{fig:supp-prim-contact} shows.
A small lead limit keeps the impact low, and the call reports the contact height corrected for the lag of the detector, the impact peak, and the stiffness when the hand pressed on.
On the real arm with a closed gripper it first re-tares from a still reading, so that the tare is taken at the pose of the contact.
\tool{probe} runs the same motion for at most an \code{amplitude} and returns to its start with \tool{move\_to}, reporting whether contact was found.

\begin{figure*}[!t]
  \centering
  \begin{minipage}[t]{\cardgfx}\vspace{0pt}\includegraphics[width=\linewidth]{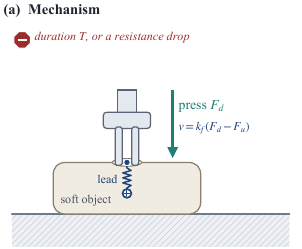}\end{minipage}\hfill
  \begin{minipage}[t]{\cardcode}\vspace{0pt}
  \begin{pcode}{(b)\enspace Pseudocode}
  \PL{0}{\Kw{input} direction $u$, force $\Fq{F_d}$, duration $T$, \code{max\_dist}, stop on a drop or not \Cm{$L=30$\,mm}}
  \PL{0}{$p_0\leftarrow p$, \Fn{Run} with, every tick,}
  \PL{1}{$\Fq{F_u}\leftarrow-\Fq{F}\cdot u$}
  \PL{1}{\Kw{if} asked \Kw{and} $t\ge0.6$\,s \Kw{and} the two-tick mean of $\Fq{F_u}$ fell by $\max(2\,\mathrm{N},\,\text{peak}/2)$ from a peak of at least 3\,N \Kw{then stop} \code{resistance\_drop}}
  \PL{1}{$v\leftarrow\mathrm{clip}\big(k_f(\Fq{F_d}-\Fq{F_u}),\pm v_{\mathrm{near}}\big)$ \Cm{\cref{eq:supp-admit}}}
  \PL{1}{\Kw{if} $(\Tq{p^{*}}-p_0)\cdot u\ge\code{max\_dist}$ \Kw{and} $v>0$ \Kw{then} $v\leftarrow0$}
  \PL{1}{$\Tq{\Delta p^{*}}\leftarrow\Tq{u}\,v\,\Delta t$, \ $\mathit{done}\leftarrow t\ge T$}
  \PL{0}{\Kw{return} mean $\Fq{F_n}\pm\sigma$ over the last 80\% of rows in contact}
  \end{pcode}
  \end{minipage}\hfill
  \begin{minipage}[t]{\cardgfx}\vspace{0pt}\includegraphics[width=\linewidth]{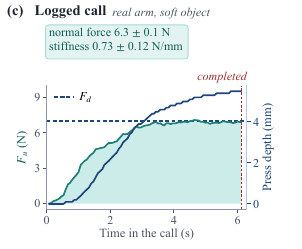}\end{minipage}
  \caption{\tool{push}. (a)~The target leads into the surface until the reaction equals $F_d$, which turns force error into target velocity. (b)~Pseudocode. (c)~Pressing a soft object on the real arm with $F_d=7$\,N for 6\,s. The force settles at the target while the hand keeps sinking, and the tool result reports the mean force and the stiffness.}
  \label{fig:supp-prim-push}
\end{figure*}

\paragraph{Holding force.}
\tool{push} regulates the reaction along $u$ to $F_d$ with the admittance law of \cref{eq:supp-admit}, as in \cref{fig:supp-prim-push}.
The target advances at most \code{max\_dist}, and on the real arm the 25\,mm lead is requested once $F_u$ reaches 5\,N.
The call ends after its duration, or, when the model asks for it, on a resistance drop, which \tool{push} detects itself on the raw force along $u$ because the generic detector smooths too much to catch a click.
It reports the mean normal force over the last 80\% of the rows in contact.

\begin{figure*}[!t]
  \centering
  \begin{minipage}[t]{\cardgfx}\vspace{0pt}\includegraphics[width=\linewidth]{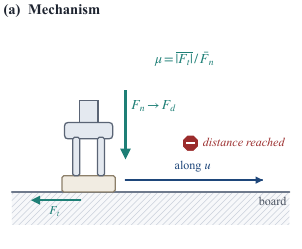}\end{minipage}\hfill
  \begin{minipage}[t]{\cardcode}\vspace{0pt}
  \begin{pcode}{(b)\enspace Pseudocode}
  \PL{0}{\Kw{input} direction $u$, \code{distance}, force $\Fq{F_d}$, normal $n$ \Cm{$L=20$\,mm}}
  \PL{0}{$u\leftarrow\mathrm{unit}\big(u-(u\cdot n)\,n\big)$, $p_0\leftarrow p$, \Fn{Run} with, every tick,}
  \PL{1}{\Kw{if} $(p-p_0)\cdot n<-2$\,cm \Kw{then} $\mathit{done}$ \Cm{left the surface}}
  \PL{1}{$v_n\leftarrow\mathrm{clip}\big(k_f(\Fq{F_d}-\Fq{F_n}),\pm v_{\mathrm{near}}\big)$}
  \PL{1}{$\mathit{done}\leftarrow(p-p_0)\cdot u\ge\code{distance}$}
  \PL{1}{$\Tq{\Delta p^{*}}\leftarrow\big(\Tq{u}\,v_{\mathrm{near}}\,[\neg\mathit{done}]+n\,v_n\big)\Delta t$}
  \PL{0}{\Kw{return} mean $\Fq{F_n}\pm\sigma$, \ $\mu=\overline{|\Fq{F_t}|}/\bar{\Fq{F}}_n\pm\sigma_\mu$, flags}
  \end{pcode}
  \end{minipage}\hfill
  \begin{minipage}[t]{\cardgfx}\vspace{0pt}\includegraphics[width=\linewidth]{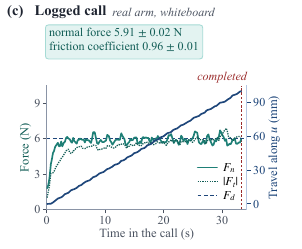}\end{minipage}
  \caption{\tool{slide}. (a)~The tool presses with $F_n$ regulated to $F_d$ while it travels along $u$, and the tangential force $F_t$ gives the friction coefficient. (b)~Pseudocode. (c)~One stroke of wiping the whiteboard on the real arm. The normal force holds at $F_d$ over the 10\,cm stroke and the tangential force is almost as large, which gives a friction coefficient near one.}
  \label{fig:supp-prim-slide}
\end{figure*}

\tool{slide} presses along $n$ by admittance while it moves at constant speed along $u$ projected onto the surface, so friction is measured rather than controlled, as \cref{fig:supp-prim-slide} shows.
The full stiffness and a 20\,mm lead let the tangential spring overcome friction, and on the real arm at most 20\,mm of the 25\,mm lead goes to the press so that some is left for the motion.
It stops after \code{distance} of measured travel or after backing 2\,cm off the surface.
It reports the mean normal force and the friction coefficient over the same window as \tool{push}, with flags \code{lost\_contact} and \code{stick\_slip}.

\begin{figure*}[!t]
  \centering
  \begin{minipage}[t]{\cardgfx}\vspace{0pt}\includegraphics[width=\linewidth]{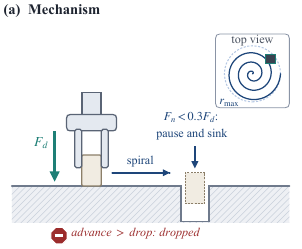}\end{minipage}\hfill
  \begin{minipage}[t]{\cardcode}\vspace{0pt}
  \begin{pcode}{(b)\enspace Pseudocode}
  \PL{0}{\Kw{input} normal $n$, force $\Fq{F_d}$, range $r_{\max}$, pitch $a$, \code{drop} \Cm{$L=8$\,mm}}
  \PL{0}{descend along $n$ until $\Fq{F_n}\ge0.3\Fq{F_d}$, then $c\leftarrow p$, $\theta\leftarrow0$}
  \PL{0}{\Fn{Run} with, every tick,}
  \PL{1}{$d\leftarrow(p-c)\cdot n$ \Cm{advance}}
  \PL{1}{\Kw{if} $d-\min_{1\,\mathrm{s}}d>\code{drop}$, $\Fq{F_n}<0.3\Fq{F_d}$, and the lateral force is small \Kw{then stop} \code{dropped}}
  \PL{1}{\Kw{if} $\Fq{F_n}<0.3\Fq{F_d}$ \Kw{then} $\Tq{\Delta p^{*}}\leftarrow1.5\,n\,v_{\mathrm{near}}\Delta t$, next tick \Cm{pause, sink}}
  \PL{1}{$v_n\leftarrow\mathrm{clip}\big(k_f(\Fq{F_d}-\Fq{F_n}),\pm v_{\mathrm{near}}\big)$}
  \PL{1}{$\Tq{\theta}\leftarrow\Tq{\theta}+v_{\mathrm{near}}\Delta t/\max(r,a/4)$, \ $r\leftarrow a\Tq{\theta}/2\pi$}
  \PL{1}{\Kw{if} $r>r_{\max}$ \Kw{then} $\mathit{done}$ \Cm{no hole found}}
  \PL{1}{$\Tq{\Delta p^{*}}\leftarrow\big[c+r(\cos\Tq{\theta}\,e_1+\sin\Tq{\theta}\,e_2)-\Tq{p^{*}}\big]_{\perp n}+n\,v_n\Delta t$}
  \PL{0}{\Kw{return} advance $d$, lateral force, flag}
  \end{pcode}
  \end{minipage}\hfill
  \begin{minipage}[t]{\cardgfx}\vspace{0pt}\includegraphics[width=\linewidth]{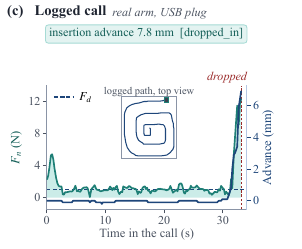}\end{minipage}
  \caption{\tool{search}. (a)~The part presses with $F_d$ and follows a spiral, and over the hole the reaction falls, the spiral pauses, and the part sinks. (b)~Pseudocode of the slide mode, with $e_1,e_2\perp n$. (c)~The call of Figure~2 of the main paper on the real arm. $F_n$ holds near $F_d=1$\,N for half a minute while the logged path spirals, then the plug drops and the thrust drives it in.}
  \label{fig:supp-prim-search}
\end{figure*}

\paragraph{Finding and seating.}
\tool{search} presses lightly and moves the target on an Archimedean spiral of pitch $a$ about its start until the part sinks, which Sec.~3.4 of the main paper walks through and \cref{fig:supp-prim-search} details.
It first descends and arms on surface contact, ending with \code{no\_surface} after 3\,cm without one.
When $F_n$ falls below $0.3F_d$, the part is over a hole, so the spiral pauses and the target falls, and a part that is already deeper than \code{drop} and blocked sideways also counts as \code{dropped}.
In simulation a lateral force above 6\,N for 0.3\,s means a corner caught on the rim, so the part lifts, moves 6\,mm along the wall's reaction, and restarts the spiral there.
On the real arm, once the advance gains over 0.25\,mm on its recent median, the target thrusts down with the 25\,mm lead toward $\max(F_d,15\,\mathrm{N})$, limited so that no force estimate would reach 23\,N within 0.3\,s, and 1.5\,mm of gain after 2\,s means \code{dropped}.
Besides \code{dropped} and \code{completed}, the call can end \code{stuck}, \code{out\_of\_range}, or \code{lost\_contact}.
A hop mode, the default in simulation, probes the spiral points one at a time instead of dragging the part, and a probe deeper than earlier ones by \code{drop} means \code{dropped}.

\begin{figure*}[!t]
  \centering
  \begin{minipage}[t]{\cardgfx}\vspace{0pt}\includegraphics[width=\linewidth]{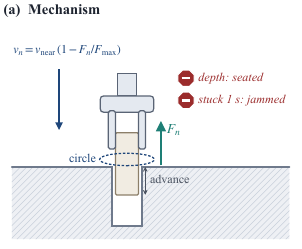}\end{minipage}\hfill
  \begin{minipage}[t]{\cardcode}\vspace{0pt}
  \begin{pcode}{(b)\enspace Pseudocode}
  \PL{0}{\Kw{input} \code{depth}, normal $n$, \code{max\_force} $\Fq{F_{\max}}$ \Cm{$L=2$\,mm, stiff along $n$}}
  \PL{0}{$p_0\leftarrow p$, \Fn{Run} with, every tick,}
  \PL{1}{$d\leftarrow(p-p_0)\cdot n$, \ \Kw{if} $d\ge\code{depth}$ \Kw{then stop} \code{seated}}
  \PL{1}{\Kw{if} $t>1$\,s \Kw{and} $d-\min_{1\,\mathrm{s}}d<1$\,mm \Kw{then stop} \code{jammed} if $\Fq{F_n}\ge0.5\Fq{F_{\max}}$, else \code{stalled}}
  \PL{1}{$v_n\leftarrow\mathrm{clip}\big(v_{\mathrm{near}}(1-\Fq{F_n}/\Fq{F_{\max}}),\pm v_{\mathrm{near}}\big)$}
  \PL{1}{$g\leftarrow p_0+w\,(\cos2\pi t\,e_1+\sin2\pi t\,e_2)$ \Cm{$w=2$\,mm, 1\,Hz}}
  \PL{1}{$\Tq{\Delta p^{*}}\leftarrow\tfrac12\big[g-\Tq{p^{*}}\big]_{\perp n}+n\,v_n\Delta t$}
  \PL{0}{\Kw{return} advance $d$, lateral force, seating flag}
  \end{pcode}
  \end{minipage}\hfill
  \begin{minipage}[t]{\cardgfx}\vspace{0pt}\includegraphics[width=\linewidth]{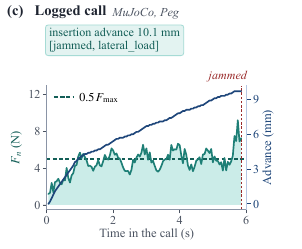}\end{minipage}
  \caption{\tool{insert}. (a)~The part advances more slowly as the reaction rises and circles slightly so that a caught corner can find the opening. (b)~Pseudocode. (c)~A call on Peg in MuJoCo. The peg advances 10\,mm while $F_n$ hovers near $0.5F_{\max}$, then progress stops and the call ends \code{jammed}.}
  \label{fig:supp-prim-insert}
\end{figure*}

\tool{insert} advances along $n$ with a speed that falls linearly with the reaction while it traces a 2\,mm circle at 1\,Hz, as in \cref{fig:supp-prim-insert}.
In MuJoCo it is stiff along $n$ and soft sideways, so the part can center itself.
It reports \code{seated} at the requested depth, and after 1\,s without 1\,mm of progress it reports \code{jammed} if $F_n\ge0.5F_{\max}$ and \code{stalled} otherwise, together with the advance, the lateral force, and a seating flag.
In the episode of Figure~2 of the main paper, the model followed a stalled \tool{insert} with \tool{wiggle}.

\begin{figure*}[!t]
  \centering
  \begin{minipage}[t]{\cardgfx}\vspace{0pt}\includegraphics[width=\linewidth]{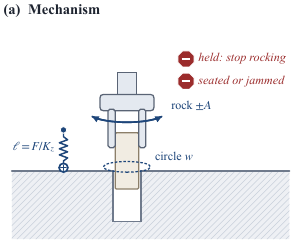}\end{minipage}\hfill
  \begin{minipage}[t]{\cardcode}\vspace{0pt}
  \begin{pcode}{(b)\enspace Pseudocode}
  \PL{0}{\Kw{input} \code{depth}, force $\Fq{F}$, yaw amplitude $A$, shift $w$, duration $T$ \Cm{real arm only}}
  \PL{0}{$p_0\leftarrow p$, phase $\leftarrow$ rock, \Fn{Run} with, every tick,}
  \PL{1}{$\Fq{F_e}\leftarrow$ ramp from $\min(8\,\mathrm{N},\Fq{F})$ to $\Fq{F}$ over 2\,s, \ $\Tq{p^{*}_z}\leftarrow p_z-\min(\Fq{F_e}/K_z,25\,\mathrm{mm})$}
  \PL{1}{\Kw{if} phase is rock \Kw{then}}
  \PL{2}{$\Tq{\psi^{*}}\leftarrow\rho A\sin(2\pi t/T_\psi)$, \ $\Tq{p^{*}_{xy}}\leftarrow p_{0,xy}+\rho w\,[\cos,\sin](2\pi t/3.9\,\mathrm{s})$}
  \PL{2}{\Kw{if} the hand stopped following the circle \Kw{or} the depth is reached \Kw{then} phase $\leftarrow$ press, freeze $\Tq{\psi^{*}},\Tq{p^{*}_{xy}}$}
  \PL{1}{\Kw{else if} the part sank under 0.2\,mm in 1\,s at full force, \Kw{or} pressed 4\,s \Kw{then stop}}
  \PL{1}{\Kw{if} $t\ge T$ \Kw{then stop}}
  \PL{0}{label \code{seated} if the depth below the surface reached \code{depth}, else \code{jammed} or \code{not\_seated}}
  \end{pcode}
  \end{minipage}\hfill
  \begin{minipage}[t]{\cardgfx}\vspace{0pt}\includegraphics[width=\linewidth]{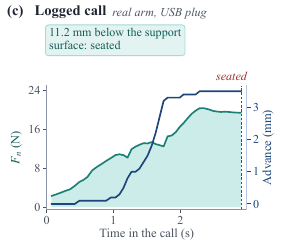}\end{minipage}
  \caption{\tool{wiggle}. (a)~A fixed lead presses the part down while the yaw rocks and the target circles, and once the part is held sideways the rocking stops. (b)~Pseudocode, with $\rho=\min(1,t/1\,\mathrm{s})$ and $T_\psi=\max(3\,\mathrm{s},2\pi A/0.126\,\mathrm{rad/s})$. (c)~The call after the stalled \tool{insert} in the episode of Figure~2 of the main paper. The press ramps up, the plug sinks 3\,mm in 3\,s, and the call ends \code{seated}.}
  \label{fig:supp-prim-wiggle}
\end{figure*}

\tool{wiggle}, which runs only on the real arm, presses the held part down with the fixed lead $\ell=\min(F/K_z,25\,\mathrm{mm})$ and rocks it so that a chamfer can guide it in, as \cref{fig:supp-prim-wiggle} shows.
The press ramps from $\min(8\,\mathrm{N},F)$ to $F$ over 2\,s, the yaw rocks with amplitude $A$, and the target circles sideways with radius $w$ and a period of 3.9\,s, so the two motions do not synchronize.
Once the hand stops following the circle, the part is held sideways, so the rocking freezes and the tool only presses until the part stops sinking.
It ends \code{seated} at sufficient depth below the support surface and \code{jammed} otherwise, and it reports the depth.

\begin{figure*}[!t]
  \centering
  \begin{minipage}[t]{\cardgfx}\vspace{0pt}\includegraphics[width=\linewidth]{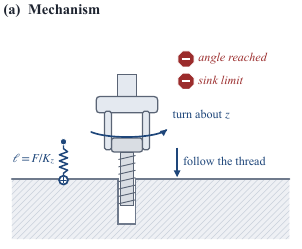}\end{minipage}\hfill
  \begin{minipage}[t]{\cardcode}\vspace{0pt}
  \begin{pcode}{(b)\enspace Pseudocode}
  \PL{0}{\Kw{input} angle $\phi$, press $\Fq{F}$ \Cm{real arm only, segments of at most 30$^{\circ}$}}
  \PL{0}{$\ell\leftarrow\min(\Fq{F}/K_z,25\,\mathrm{mm})$, $p_0\leftarrow p$, \Fn{Run} with, every tick,}
  \PL{1}{\Kw{if} the tip sank more than 10\,mm \Kw{then stop} \code{sink\_limit}}
  \PL{1}{$\Tq{p^{*}_z}\leftarrow p_z-\ell$, at most 8\,mm/s pressing and $v_{\mathrm{near}}$ following \Cm{press, follow}}
  \PL{1}{\Kw{if} pressing \Kw{then} $\Tq{p^{*}_{xy}}\leftarrow p_{0,xy}$, and turn once the tip stops sinking or after 6\,s}
  \PL{1}{\Kw{else} $\Tq{\psi^{*}}\leftarrow\Tq{\psi^{*}}+\mathrm{sgn}(\phi)\,5^{\circ}/\mathrm{s}\cdot\Delta t$, \ $\Tq{p^{*}_{xy}}\leftarrow p_{0,xy}+0.3\,(p-p_0)_{xy}$}
  \PL{1}{$\mathit{done}$ once $\phi$ is turned}
  \PL{0}{\Kw{return} advance while pressing and turning, mm per turn, trend label}
  \end{pcode}
  \end{minipage}\hfill
  \begin{minipage}[t]{\cardgfx}\vspace{0pt}\includegraphics[width=\linewidth]{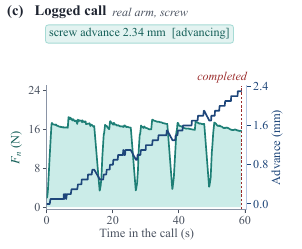}\end{minipage}
  \caption{\tool{screw}. (a)~A fixed lead presses the screw into the thread while the hand turns about the vertical and follows the thread down. (b)~Pseudocode. (c)~A $-150^{\circ}$ call on the real arm, run as five segments. Each segment re-presses, the force dips between them, and the advance climbs in steps as the thread pulls the screw in.}
  \label{fig:supp-prim-screw}
\end{figure*}

\paragraph{Turning under load.}
\tool{screw}, which runs only on the real arm, first presses with the fixed lead $\ell=\min(F/K_z,25\,\mathrm{mm})$ and then turns about the vertical at 5$^{\circ}$/s while it keeps pressing and follows the thread down, as \cref{fig:supp-prim-screw} shows.
The turn starts once the tip stops sinking under a full lead, or after 6\,s.
While it turns, the lateral target becomes $p_0+0.3\,(p-p_0)$, and a call is split into segments of at most 30$^{\circ}$.
It ends \code{completed} at the angle or \code{sink\_limit} after 10\,mm per segment, and it reports the advance while pressing and while turning, millimeters per turn, and \code{advancing}, \code{rising}, or \code{not\_advancing}.

\begin{figure*}[!t]
  \centering
  \begin{minipage}[t]{\cardgfx}\vspace{0pt}\includegraphics[width=\linewidth]{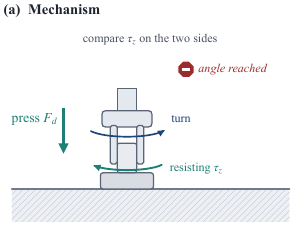}\end{minipage}\hfill
  \begin{minipage}[t]{\cardcode}\vspace{0pt}
  \begin{pcode}{(b)\enspace Pseudocode}
  \PL{0}{\Kw{input} angle $\phi$, force $\Fq{F_d}$, rate $\omega$, \code{press\_time} $T_p$ \Cm{ForceBench only, $L=30$\,mm}}
  \PL{0}{$p_0\leftarrow p$, \Fn{Run} with, every tick,}
  \PL{1}{$v\leftarrow\mathrm{clip}\big(k_f(\Fq{F_d}-\Fq{F_z}),\pm v_{\mathrm{near}}\big)$, \ $\Tq{\Delta p^{*}}\leftarrow-\hat z\,v\,\Delta t$ \Cm{hold the press}}
  \PL{1}{\Kw{if} $t\ge T_p$ \Kw{then}}
  \PL{2}{$\Tq{\psi^{*}}\leftarrow\Tq{\psi^{*}}+\mathrm{sgn}(\phi)\,\omega\Delta t$ until $|\phi|$ is turned}
  \PL{2}{$\Fq{\tau_{\mathrm{res}}}\leftarrow\mathrm{sgn}(\phi)\,\Fq{\tau_z}$, added to the side's samples while moving 8 to 25$^{\circ}$ out}
  \PL{1}{$\mathit{done}$ once $\phi$ is turned \Cm{or the optional \code{stop\_depth}}}
  \PL{0}{\Kw{return} press force, depth, mean $\Fq{\tau_{\mathrm{res}}}$ over the second half, peak, side medians}
  \end{pcode}
  \end{minipage}\hfill
  \begin{minipage}[t]{\cardgfx}\vspace{0pt}\includegraphics[width=\linewidth]{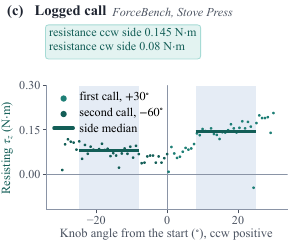}\end{minipage}
  \caption{\tool{press\_turn}. (a)~The hand holds the knob down with $F_d$ while it turns, and the resisting torque is read on each side of the start. (b)~Pseudocode. (c)~The two calls of one successful Stove Press episode of GPT-6.1 Sol in ForceBench, $+30^{\circ}$ and then $-60^{\circ}$. The knob resists less on the clockwise side, and the tool result reports the median resistance of each side.}
  \label{fig:supp-prim-pressturn}
\end{figure*}

\tool{press\_turn}, which exists only in ForceBench, holds a press along $-z$ by \cref{eq:supp-admit} and, after \code{press\_time}, also turns about the vertical, on Stove Press and Nut Thread.
It exists because each call resets $p^{*}=p$, which drops a \tool{push} press before the next \tool{rotate\_z}.
Besides the press force and depth and the resisting torque over the second half of the turn, it reports on Stove Press the median resistance on each side of the start, which \cref{fig:supp-prim-pressturn} shows.
The no-force control uses \tool{move\_turn} instead, which Nut Thread also offers with force, and which moves the target down by a set amount and turns while it follows a commanded height, and whose description tells the model that the arm acts as a spring of about 1\,N per mm, so that the control lacks force rather than a motion.

\begin{figure*}[!t]
  \centering
  \begin{minipage}[t]{\cardgfx}\vspace{0pt}\includegraphics[width=\linewidth]{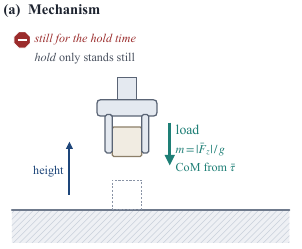}\end{minipage}\hfill
  \begin{minipage}[t]{\cardcode}\vspace{0pt}
  \begin{pcode}{(b)\enspace Pseudocode}
  \PL{0}{\Kw{input} \code{height}, \code{hold\_time} $T_h$ \Cm{no contact events while holding}}
  \PL{0}{$p_0\leftarrow p$, \Fn{Run} with, every tick,}
  \PL{1}{\Kw{if} $p_z-p_{0,z}<\code{height}$ \Kw{and} not yet still \Kw{then} $\Tq{\Delta p^{*}}\leftarrow\hat z\,v_{\mathrm{near}}\Delta t$}
  \PL{1}{\Kw{else} stand still, \ $\mathit{done}$ after $T_h$}
  \PL{0}{$W\leftarrow$ the last $\max(5,n/4)$ rows \Cm{the still window}}
  \PL{0}{$m\leftarrow|\bar{\Fq{F}}_z|/g\pm\sigma_m$, \ $\Fq{b_c}\leftarrow\bar{\Fq{F}}$ over $W$ \Cm{Eq.~4 of the main paper}}
  \PL{0}{\Kw{return} $m$ and the center-of-mass offset from $\bar{\Fq{\tau}}_x,\bar{\Fq{\tau}}_y$}
  \PL{0}{\Fn{Hold}$(T)$: \Fn{Run} with $\Tq{\Delta p^{*}}=0$ until $t\ge T$}
  \end{pcode}
  \end{minipage}\hfill
  \begin{minipage}[t]{\cardgfx}\vspace{0pt}\includegraphics[width=\linewidth]{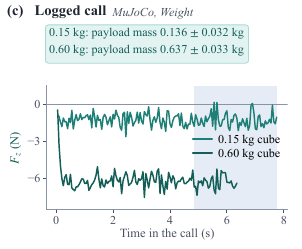}\end{minipage}
  \caption{\tool{lift} and \tool{hold}. (a)~The hand lifts the grasped load and holds still, so the vertical force is the weight. (b)~Pseudocode. (c)~Two lifts on Weight in MuJoCo, for a light and a heavy cube on two seeds. The shaded still windows give masses within about one $\sigma$ of 0.15 and 0.60\,kg.}
  \label{fig:supp-prim-lift}
\end{figure*}

\paragraph{Measuring.}
\tool{lift} rises by \code{height} at $v_{\mathrm{near}}$ and then stands still for \code{hold\_time}, so that the force on the hand is the load, as \cref{fig:supp-prim-lift} shows, and \tool{hold} only stands still.
In simulation both report the payload mass of Eq.~4 of the main paper and the center-of-mass offset over the still rows, and \tool{lift} sets the held-load tare from them.
On the real arm the runner instead gives the still reading, its change since the grasp, and a weight in grams, which serves to compare loads within an episode.

\begin{figure*}[!t]
  \centering
  \begin{minipage}[t]{\cardgfx}\vspace{0pt}\includegraphics[width=\linewidth]{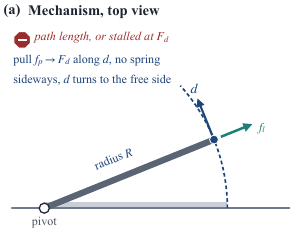}\end{minipage}\hfill
  \begin{minipage}[t]{\cardcode}\vspace{0pt}
  \begin{pcode}{(b)\enspace Pseudocode}
  \PL{0}{\Kw{input} direction $d$, force $\Fq{F_d}$, \code{distance} \Cm{$L=30$\,mm, soft in the horizontal plane}}
  \PL{0}{$d\leftarrow$ horizontal unit vector, \Fn{Run} with, every tick,}
  \PL{1}{$l\leftarrow\hat z\times d$, \ $\Fq{f_p}\leftarrow-\Fq{F}\cdot d$, \ $\Fq{f_l}\leftarrow\Fq{F}\cdot l$}
  \PL{1}{\Kw{if} path $\ge5$\,mm \Kw{and} $\Fq{f_p}\ge\max(1\,\mathrm{N},\Fq{F_d}/4)$ \Kw{then} turn $d$ by $0.1\,\mathrm{atan2}(\Fq{f_l},\Fq{f_p})$, at most $\min(3^{\circ},\text{step}/0.15\,\mathrm{m})$}
  \PL{1}{$\Tq{\Delta p^{*}}\leftarrow\big(d\,c(k_f(\Fq{F_d}-\Fq{f_p}))+l\,c(k_f\Fq{f_l})\big)\Delta t+l\,\big((p-\Tq{p^{*}})\cdot l\big)$ \Cm{$c$ clips to the speed}}
  \PL{1}{$\mathit{done}$ at the path length, \ \code{stalled} if still for 1\,s with $\bar{\Fq{f}}_p\ge0.95\Fq{F_d}$}
  \PL{0}{\Kw{return} $\Fq{f_p}\pm\sigma$, lateral force, path turn, radius and pivot of a circle fit}
  \end{pcode}
  \end{minipage}\hfill
  \begin{minipage}[t]{\cardgfx}\vspace{0pt}\includegraphics[width=\linewidth]{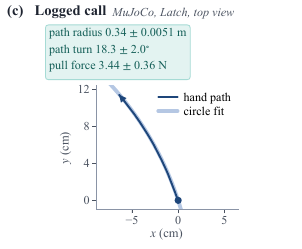}\end{minipage}
  \caption{\tool{pull}. (a)~Seen from above, the hand pulls along $d$ with no sideways spring, so the door's constraint bends the path into an arc and $d$ follows it. (b)~Pseudocode. (c)~The pull that opened the door on Latch in MuJoCo. The logged hand path turns by 18$^{\circ}$ and lies on the fitted circle.}
  \label{fig:supp-prim-pull}
\end{figure*}

\tool{pull} moves a gripped handle along the path that its mechanism allows, on Latch, Lock, and Drawer Pull, using force only, as \cref{fig:supp-prim-pull} shows.
It regulates the pull force along $d$ by admittance and lets the hand follow any sideways force, so no spring acts sideways, and after 5\,mm of travel it turns $d$ toward the free side by at most 3$^{\circ}$ per tick.
It ends \code{completed} at the path length, \code{stalled} when the hand stops while $f_p\ge0.95F_d$, or \code{joint\_limit}.
It reports the pull and lateral forces, the turn of the path with the radius and pivot of a circle fit, the free direction, and \code{mechanism\_stopped}.
On the real arm a direction with a vertical part runs a straight pull with the wrist fixed.

\begin{figure*}[!t]
  \centering
  \begin{minipage}[t]{\cardgfx}\vspace{0pt}\includegraphics[width=\linewidth]{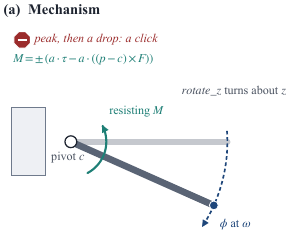}\end{minipage}\hfill
  \begin{minipage}[t]{\cardcode}\vspace{0pt}
  \begin{pcode}{(b)\enspace Pseudocode}
  \PL{0}{\Kw{input} axis $a$, pivot $c$, angle $\phi_{\max}$, rate $\omega$, stop on a drop or not}
  \PL{0}{$p_0\leftarrow p$, $q_0\leftarrow q^{*}$, $\phi\leftarrow0$, \Fn{Run} with, every tick,}
  \PL{1}{$\Tq{\phi}\leftarrow\Tq{\phi}+\mathrm{sgn}(\phi_{\max})\min(\omega\Delta t,\,|\phi_{\max}|-|\Tq{\phi}|)$}
  \PL{1}{$\Tq{p^{*}}\leftarrow c+\mathrm{Rot}(a,\Tq{\phi})(p_0-c)$, \ $\Tq{q^{*}}\leftarrow\mathrm{Rot}(a,\Tq{\phi})\,q_0$}
  \PL{1}{$\Fq{M}\leftarrow\mathrm{sgn}(\phi_{\max})\big(a\cdot\Fq{\tau}-a\cdot((p-c)\times\Fq{F})\big)$, averaged over 4 ticks}
  \PL{1}{after 0.6\,s, a fall of $\max(0.15\,\mathrm{N\,m},\,0.2\,\text{peak})$ below the peak on 2 ticks is a click}
  \PL{1}{\Kw{if} asked \Kw{and} a click \Kw{then stop} \code{resistance\_drop}, \ $\mathit{done}$ once $\phi_{\max}$ is turned}
  \PL{0}{\Kw{return} turned angle, radius, and in ForceBench $\Fq{M}$, its peak, its end value}
  \PL{0}{\Fn{Rotate\_z}$(\phi)$: turn about the vertical through $p$ with $\Tq{p^{*}}$ fixed, within the wrist range}
  \end{pcode}
  \end{minipage}\hfill
  \begin{minipage}[t]{\cardgfx}\vspace{0pt}\includegraphics[width=\linewidth]{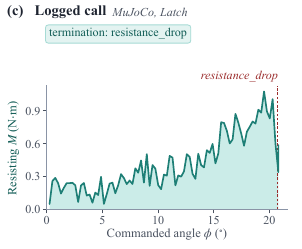}\end{minipage}
  \caption{\tool{turn} and \tool{rotate\_z}. (a)~The hand turns a lever about its pivot, and a peak of the resisting moment followed by a drop marks the latch giving way. (b)~Pseudocode. (c)~Pressing the latch handle on Latch in MuJoCo, recomputed from the logged rows. The moment rises to about 1\,N$\cdot$m and falls at 21$^{\circ}$, where the call stops on \code{resistance\_drop} and the model goes on to pull.}
  \label{fig:supp-prim-turn}
\end{figure*}

\paragraph{Turning.}
\tool{turn} rotates the hand rigidly about an axis $a$ through a pivot $c$, so the tool center moves on a circle and the tool orientation turns with it, on Latch, the Faucet tasks, and the real arm, as \cref{fig:supp-prim-turn} shows.
With force, the model can ask it to stop on a resistance drop, which it detects on the resisting moment about the turn axis in simulation and, without a verified torque sign, on the force along the arc tangent on the real arm.
ForceBench can also print the moment, its peak, and its end value, which was off in all reported runs.
\tool{rotate\_z} turns the tool about the vertical with the position target fixed, and the wrist joint is kept within $\pm2.6$\,rad in MuJoCo and $-0.8$ to 2.4\,rad on the real arm. It reports the yaw change and whether the angle was clipped.

\begin{figure*}[!t]
  \centering
  \begin{minipage}[t]{\cardgfx}\vspace{0pt}\includegraphics[width=\linewidth]{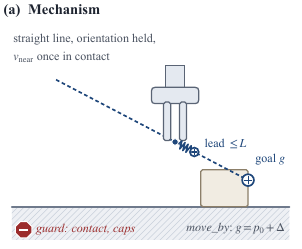}\end{minipage}\hfill
  \begin{minipage}[t]{\cardcode}\vspace{0pt}
  \begin{pcode}{(b)\enspace Pseudocode}
  \PL{0}{\Kw{input} goal $g$, speed $v$ \Cm{$L=30$\,mm, 12\,mm if guarded, 8\,mm on the real arm}}
  \PL{0}{\Fn{Run} with stops $S=\{\code{contact}\}$ and, every tick,}
  \PL{1}{\Kw{if} $\lVert g-p\rVert<$ tolerance \Kw{then} $\mathit{done}$}
  \PL{1}{$s\leftarrow v_{\mathrm{near}}$ \Kw{if} in contact \Kw{else} $v$}
  \PL{1}{$\Tq{\Delta p^{*}}\leftarrow\frac{g-\Tq{p^{*}}}{\lVert g-\Tq{p^{*}}\rVert}\min\big(\lVert g-\Tq{p^{*}}\rVert,\,s\,\Delta t\big)$}
  \PL{0}{\Kw{return} kinematics, and a note if the call ended over 5\,mm short}
  \PL{0}{\Fn{Move\_by}$(\Delta)$: \Fn{Move\_to}$(p_0+\Delta)$}
  \end{pcode}
  \end{minipage}\hfill
  \begin{minipage}[t]{\cardgfx}\vspace{0pt}\includegraphics[width=\linewidth]{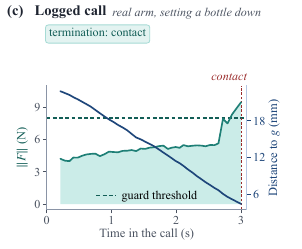}\end{minipage}
  \caption{\tool{move\_to} and \tool{move\_by}. (a)~The target runs ahead of the hand on a straight line, and the guard stops the move on contact. (b)~Pseudocode. (c)~Setting a held bottle down on the real arm. The hand stops 4.5\,mm short of its goal when the force passes the guard threshold of 8\,N.}
  \label{fig:supp-prim-move}
\end{figure*}

\paragraph{Moving by position.}
\tool{move\_to} moves the tool center on a straight line to a goal $g$ with the orientation unchanged, and \tool{move\_by} does the same toward $p_0+[dx,dy,dz]$, as \cref{fig:supp-prim-move} shows.
In simulation the speed drops to $v_{\mathrm{near}}$ in contact.
On the real arm joint friction can stop the hand short of $g$, so a stuck hand near $g$ gets a target bias that grows up to the 8\,mm lead and retracts within 1\,mm of $g$.
Both stop on contact at $f_c$, or at the collision limits on the real arm, where they can also end \code{stalled\_short}.
They report kinematics and a note if the call ended over 5\,mm short.
\tool{point\_tool} turns the fingers toward a direction with the tool center fixed, on Latch, Lock, Drawer Pull, and the real arm. In simulation the force zero is converted every tick as the tool turns.

\subsection{The remaining tools, guard, and validation}
\label{sec:supp-prim-constants}

The gripper and perception tools do not regulate force, and \cref{tab:supp-prim-other} summarizes them.
\tool{locate} maps a pixel to a 3D point in ForceBench and on the real arm, while MuJoCo gives object poses as text, and \tool{locate\_object} is an extra tool added for Nut Thread beyond the 22 base tools, as \cref{sec:supp-forcebench} notes.
On the real arm a gripper width outside the expected range after \tool{grasp} or \tool{release} latches a fault, after which the gripper is never opened.

\begin{table}[t]
  \caption{Tools that neither move by force nor hold a force.}
  \label{tab:supp-prim-other}
  \centering
  \footnotesize
  \setlength{\tabcolsep}{3pt}
  \begin{tabular}{@{}p{0.2\linewidth}p{0.44\linewidth}p{0.29\linewidth}@{}}
    \toprule
    Tool & What it does & What it returns \\
    \midrule
    \tool{grasp}, \tool{release} & close or open the gripper, clear the tare & opening, holding or not \\
    \tool{locate} & map a pixel to a 3D point, no motion & point, valid pixels \\
    \tool{locate\_object} & fit a known part's axis to a clicked mask & axis, overlap score \\
    \tool{scene} & describe the scene, no motion & poses, or photos \\
    \tool{done} & end the episode & logged claim only \\
    \bottomrule
  \end{tabular}
\end{table}

Beneath all tools, a fixed guard on the real-time computer enforces limits that the model cannot change.
It caps the raw reading at 25\,N and 3\,N$\cdot$m and keeps the hand inside a workspace box, and it limits the speed to 8\,mm/s, the tracking error to 15\,mm, and the age of a command to 300\,ms.
The guard clips the target speed, and a breach of any other limit ends the episode.

A call that fails validation does not move the robot, and its tool result gives the reason.
Simulation checks tool and argument names, clips force targets at 20\,N inside the laws, and allows several calls per turn.
The real arm also checks argument types, the gripper state, and reach along the path on a kinematic model, and it caps force targets at 10 to 20\,N depending on the tool and task.
It allows one call per turn, and a rejected call still uses up an action.

\section{Physics translation layer}
\label{sec:supp-physics}

This section specifies the physics translation layer of Sec.~3.5 of the main paper, which \bench, ForceBench, and the FR3 share, so that each number in a tool result can be recomputed from the primitive's log.

\subsection{Common recipe}
\label{sec:supp-phys-recipe}

Five estimators, whose constants per backend are in \cref{tab:supp-phys-constants}, turn the log of one primitive into the quantities of \cref{sec:supp-phys-quantities} by the recipe of Sec.~3.5 of the main paper.
The log holds the force, torque, and tool position of every tick and is downsampled uniformly beyond 160 rows.
Each estimator removes the zero and any held load, $F_i = F_i^{\mathrm{raw}} - c$, where $F^{\mathrm{raw}}$ is the zeroed force and the tare $c$ is set by \tool{lift} and cleared by \tool{grasp} and \tool{release}.
Payload mass and the FR3 \tool{pull} skip this tare, since the load or the resistance is their signal.
Within the window where it has one meaning, a quantity is by default a mean with its standard error, while payload mass and pull force add the residual zero bias $\sigma_b$, a ratio propagates relative errors, and a slope takes its least-squares error,
\begin{align}
  \bar{x} &= \tfrac{1}{n}\textstyle\sum_{i\in W} x_i, \qquad \sigma = s/\sqrt{n}, \label{eq:supp-phys-mean}\\
  \sigma_m &= \tfrac{1}{g}\sqrt{s_z^2/n + \sigma_b^2}, \quad \sigma_{\mathrm{pull}} = s/\sqrt{n} + \sigma_b, \label{eq:supp-phys-bias}\\
  \sigma_\mu^2/\mu^2 &= \sigma_t^2/\overline{\lvert F_t \rvert}^2 + \sigma_n^2/\bar{F}_n^2, \label{eq:supp-phys-ratio}\\
  \sigma_k &= \sqrt{\textstyle\sum_i e_i^2/(n-2)} \,\Big/ \sqrt{\textstyle\sum_i (\delta_i-\bar{\delta})^2}, \label{eq:supp-phys-slope}
\end{align}
where $W$ is the window of $n$ rows and $s$ the sample standard deviation.
$\sigma_t$ and $\sigma_n$ are the standard errors of the two means, $e_i$ the fit residuals, and $\delta_i$ the penetrations.
The flags, given with each quantity in \cref{sec:supp-phys-quantities}, are class bands fixed across tasks, warnings, and outcomes translated from how the primitive ended, and no step holds a task threshold.

\begin{table}[b]
  \caption{Estimator constants per backend. MuJoCo adds this noise and bias to its sensor, and ForceBench uses the same defaults, with its lag and contact $\sigma$ computed by formula. The other constants are in \cref{tab:supp-prim-constants}.}
  \label{tab:supp-phys-constants}
  \centering
  \footnotesize
  \setlength{\tabcolsep}{3pt}
  \let\rr\raggedright
  \begin{tabular}{@{}p{0.33\linewidth}p{0.15\linewidth}p{0.25\linewidth}p{0.19\linewidth}@{}}
    \toprule
    Constant & MuJoCo & ForceBench & FR3 \tabularnewline
    \midrule
    \rr Detector lag $\ell$ & \rr 4\,mm & \rr 5.3\,mm & \rr 0.2\,mm \tabularnewline
    \rr Contact $\sigma$ & \rr 3\,mm & \rr 4\,mm & \rr 0.3\,mm \tabularnewline
    \rr Noise per tick $\sigma_{\mathrm{noise}}$ & \rr 0.5\,N & \rr 0.5\,N & \rr 0.0688\,N \tabularnewline
    \rr Bias per episode $\sigma_b$ & \rr 0.3\,N & \rr 0.3\,N & \rr 0.0747\,N \tabularnewline
    \bottomrule
  \end{tabular}
\end{table}

\subsection{Every quantity}
\label{sec:supp-phys-quantities}

This subsection gives every quantity that a tool result can carry and its flags, grouped by the primitives that report it.
Each paragraph states the window and the $\sigma$ of a quantity.
Here $u$ is the call's direction, or else its start-to-end direction, and $\hat{n}$ its normal.
The tool position $p$ starts at $p_0$, $t_c$ is the time of contact, and $F$ is tared unless marked raw.
Endings that no rule below translates, such as \code{stuck} or \code{timeout}, pass through by name.

\paragraph{Held load.}
\tool{lift} and \tool{hold} report the payload mass $m=\lvert\bar{F}^{\mathrm{raw}}_z\rvert/g$ with the gripper closed and $g=9.81\,\mathrm{m/s^2}$, averaged over the last \code{hold\_time} or else the last half of the call, with $\sigma_m$ of \eqref{eq:supp-phys-bias}.
When $m\ge 0.03$\,kg and the load hangs, they also report the center-of-mass offset over the same window, printed as $\lvert r\rvert$ and the nearest base axis.
A load of weight $mg$ at horizontal lever $r$ gives the torque $(-mg\,r_y,\ mg\,r_x,\ 0)$, and a sign constant $s_\tau$ set on the simulator's true lever maps the sensor's raw torque rows to $r$,
\begin{align}
  r_x &= s_\tau\frac{\bar{T}_y}{mg}, \quad r_y = -s_\tau\frac{\bar{T}_x}{mg}, \quad s_\tau=-1, \label{eq:supp-phys-lever}\\
  \sigma_r &= \sqrt{(\sigma_\tau/mg)^2 + (|r|\,\sigma_m/m)^2}, \label{eq:supp-phys-comsigma}
\end{align}
with $\sigma_\tau = \sqrt{(s_x^2+s_y^2)/n} + \tau_b$, where $s_x$ and $s_y$ are the standard deviations of the two torque axes and $\tau_b=0.01$\,N\,m is the torque bias.
The lever is taken about the sensor's vertical axis.
Both quantities appear in MuJoCo and ForceBench.
A reading is flagged \code{not\_settled} if the height varies over 3\,mm and \code{unsteady} if $s_z>3\sigma_{\mathrm{noise}}+0.5$\,N.
The mass is flagged \code{no\_load} if $m<2\sigma_m$ and \code{pushing\_up\_not\_hanging} if $\bar{F}_z>0$ without \code{no\_load}, and the offset is flagged \code{within\_noise} if $\lvert r\rvert<2\sigma_r$.
The FR3 reports weight from the still reading of \cref{sec:supp-sensing} instead of payload mass, which would use one zero point for every pose, and it omits the offset.

\paragraph{Contact.}
A call whose contact event passes the contact gate reports four quantities on all three backends.
The gate admits an event outside \tool{grasp}, \tool{release}, and \tool{pull} that lasts at least 0.15\,s and peaks at or above $0.8f_c$ within 0.3\,s, where $f_c$ is the contact threshold.
The contact height is $z_c=z(t_c)-u_z\ell$ if $\lvert u_z\rvert>0.7$, and otherwise the contact point is $s_c=(p(t_c)-u\ell-p_0)\cdot u$, both taken at the tick of $t_c$ with $\sigma=\ell/2+v\Delta t$, where $\ell$ is the detector lag of \cref{tab:supp-phys-constants}.
The contact normal $\bar{F}/\lvert\bar{F}\rvert$ and its tilt $\arccos(\lvert\bar{F}_z\rvert/\lvert\bar{F}\rvert)$ use the 6 rows from $t_c$, and the impact peak is $\max\lvert F\rvert$ from $t_c$ to $t_c+0.3$\,s, both without $\sigma$.
The stiffness $k$ is a least-squares fit $F_u=k\delta+b$ with $\delta=(p-p(t_c))\cdot u$ and $F_u=-F\cdot u$, on at least 4 of the 12 rows from $t_c$ that have $\delta>0.5$\,mm, and it is reported if $k>0.05$\,N/mm with $\sigma_k$ of \eqref{eq:supp-phys-slope}.
The impact is labeled \code{gentle} below 5\,N, \code{moderate} below 15\,N, and \code{hard} above, and the stiffness \code{soft} below 1\,N/mm, \code{firm} below 10\,N/mm, and \code{rigid} above.

\paragraph{Pressing and sliding.}
\tool{push} and \tool{slide} report the normal force, the mean of $F_n=-F\cdot u$, or of $-F\cdot\hat{n}$ on \tool{slide}, over the rows of the last 80\% of the call with $F_n>0.5f_c$, with $\sigma$ of \eqref{eq:supp-phys-mean}.
On the same rows \tool{slide} adds the friction coefficient $\mu=\overline{\lvert F_t\rvert}/\bar{F}_n$ with $F_t=F\cdot u$ and $\sigma_\mu$ of \eqref{eq:supp-phys-ratio}, and the stick-slip index $\mathrm{std}\lvert F_t\rvert/\overline{\lvert F_t\rvert}$.
All three backends report them.
If fewer than 3 rows qualify, the result says \code{lost\_contact} and gives the mean of all rows with no $\sigma$ and no friction line, and an index above 0.5 sets \code{stick\_slip}.

\paragraph{Insertion.}
\tool{search} and \tool{insert} report the advance $d=(p_{\mathrm{end}}-p_0)\cdot\hat{n}$ over the whole call and the lateral force $\lvert\overline{F-(F\cdot\hat{n})\hat{n}}\rvert$ over its last 5 rows, without $\sigma$, on all three backends.
\tool{search} ends \code{dropped\_in} on a drop or a resistance drop, \code{no\_hole\_found} at full radius, and \code{jammed} if stalled.
\tool{insert} ends \code{seated} at depth, and after under 1\,mm of advance in 1\,s it ends \code{jammed} with $F_n$ at least half of \code{max\_force} or \code{stalled\_no\_progress} with less, and otherwise \code{not\_seated}.
Both give \code{jammed\_hard} at the force cap and \code{lateral\_load} if the lateral force at the end exceeds $f_c$.

\paragraph{Pulling.}
\tool{pull} reports the pull force, the mean of $-F\cdot d_i$ along the pull direction $d_i$ of each tick, and the mean sideways force over the last 80\%, with $\sigma$ of \eqref{eq:supp-phys-bias}, or on the FR3 the mean of $-F^{\mathrm{raw}}\cdot u$.
After 3\,cm of travel it adds the path turn $\mathrm{atan2}(q_1\times q_2,\,q_1\cdot q_2)$ between the first and last 1.5\,cm of path $q_1$ and $q_2$, counter-clockwise positive, the radius and pivot of an algebraic circle fit if $\lvert\mathrm{turn}\rvert\ge 4^\circ$, and the free direction as the heading of $q_2$.
They carry a $\sigma$ of $2^\circ$, or the fit RMS plus 5\,mm for the fit, in MuJoCo and ForceBench.
The call ends \code{mechanism\_stopped} when it stalls, under 3\,mm in 1\,s with a mean pull force over 0.5\,s of at least 95\% of target, or on the FR3 under 0.2\,mm in 1\,s with a mean resistance of at least 95\% of target, and a turn under $3^\circ$ is flagged \code{straight\_line}.

\paragraph{Turning on ForceBench.}
The quantities of this paragraph and the next appear only under \paper.
On ForceBench, $\theta$ is the commanded turn about the axis $a$ through the pivot $p_v$, and $T$ is the torque row.
\tool{press\_turn} reports the resisting torque $\tau=\mathrm{sgn}(\theta)T_z$, positive against the turn, as its mean over the second half with $\sigma=s/\sqrt{n}$ and its peak over the turn.
On Stove Press it reports instead the resistance per side, the median of $\tau$ on each side of the start angle over rows moving away from it, 8 to 25$^\circ$ out and since the last grasp, from 4 rows on, with $\sigma=1.253s/\sqrt{n}$.
It also reports the press force, the mean upward reaction while turning, and the depth, the start minus the end tool height, both without $\sigma$.
\tool{turn} can also report the resisting moment $M=\mathrm{sgn}(\theta)\big[a\cdot T-a\cdot((p-p_v)\times F)\big]$, smoothed over 4 rows, as its mean over the second half, its peak over the call, and its value at the last row, without $\sigma$, but this was off in all reported runs.

\paragraph{On the FR3 only.}
Every call reports the still reading of \cref{sec:supp-sensing} and, while the hand holds a load, the weight $\hat{m}=-\Delta F_z/g$ above 20\,g, with $\Delta F$ the change since \tool{grasp} ended.
The reading takes 0.5\,s after a 0.8\,s relax, nominally, and carries a fixed $\pm 400$\,g on grams.
Every call also reports the net force peak $\max\lvert F\rvert$ over the whole call.
\tool{screw} reports the advance $\Delta z=z_{\mathrm{start}}-z_{\mathrm{end}}$, split into the sink while pressing and while turning, the advance per turn $\Delta z_{\mathrm{turn}}/\lvert\theta\rvert\times 360^\circ$ if $\lvert\theta\rvert\ge 10^\circ$, and as side load the size of the untared change in $\bar{F}_{xy}$ between its first 3 and last 5 rows, without $\sigma$.
It ends \code{advancing}, \code{rising}, or \code{not\_advancing} for a sink while turning above 0.2\,mm, below $-0.2$\,mm, or in between.
The FR3 prints no quantity when the force calibration is not valid for the pose.

\subsection{Rendering levels}
\label{sec:supp-phys-render}

L1 prints each value with its unit and any fixed geometry, and L2, the default of \paper, adds $\pm\,\sigma$ and a line with the window and flags, as \cref{fig:supp-prompts-forms} shows.

\subsection{Accuracy}
\label{sec:supp-phys-accuracy}
\label{sec:supp-phys-limits}

We check the estimators against the ground truth of the simulators.
On \bench test episodes, the mass error against simulator truth has a mean of 0.004\,kg and a standard deviation of 0.03\,kg, which matches the printed $\sigma$, and the direction cosine of the center-of-mass offset is at least 0.97.
On ForceBench, masses of 0.06 and 0.18\,kg read 0.058 and 0.177\,kg.
The printed $\sigma$ covers measurement noise only.

\section{The \texorpdfstring{\bench}{HiddenPhys} benchmark}
\label{sec:supp-benchmark}

\bench is the benchmark behind Table 5 of the main paper, and \cref{sec:supp-results} extends its results.
Everything in this section and in \cref{sec:supp-baselines} belongs to its MuJoCo backend.

\subsection{Simulation setup}
\label{sec:supp-simsetup}

\Cref{tab:supp-simconst} lists the constants of the simulation, and a few behaviors of the simulator decide what each row can sense.
Tool results omit durations, simulated time, and the finger opening while the gripper is closed, because these reveal the load.

\begin{table}[!htb]
  \caption{Constants of \bench, shared by all tasks unless stated. Noise and bias are standard deviations of Gaussians.}
  \label{tab:supp-simconst}
  \centering
  \footnotesize
  \setlength{\tabcolsep}{3pt}
  \let\rr\raggedright
  \begin{tabular}{@{}lp{0.69\linewidth}@{}}
    \toprule
    Item & Value \tabularnewline
    \midrule
    Simulator & \rr robosuite 1.5.1 on MuJoCo 3.2.7 \tabularnewline
    Robot & \rr Panda, BASIC composite controller, fixed impedance, damping ratio 1 \tabularnewline
    Action & \rr operational-space pose delta and gripper at 20\,Hz, full scale 5\,cm and 0.5\,rad per tick, no gripper on Wipe \tabularnewline
    Stiffness kp & \rr 150, 40 along a pressed axis, 10 on wrist yaw in \tool{pull}, 1500 for the stiff arm of the Lock checks \tabularnewline
    Arrival tolerance & \rr 3\,mm, 0.1\,mm on Lock \tabularnewline
    Force sensor & \rr MuJoCo wrist sensor, rotated to the world frame \tabularnewline
    Force noise & \rr 0.5\,N and 0.05\,N$\cdot$m per axis and tick \tabularnewline
    Residual bias & \rr 0.3\,N and 0.03\,N$\cdot$m per axis and episode \tabularnewline
    Zero point & \rr mean of 10 ticks at rest, at reset and after leveling the tool to within 0.5$^\circ$ \tabularnewline
    Scene noise & \rr offset per object on the true pose, fixed within an episode, 4\,mm in $x$ and $y$, 15\,mm on Peg, 2$^\circ$ in yaw \tabularnewline
    Contact, stop & \rr contact at 3\,N, stop at 25\,N or 3\,N$\cdot$m, force targets at most 20\,N \tabularnewline
    Episode budget & \rr 40 turns, 80 tool calls, 300\,s simulated, 1200\,s wall clock \tabularnewline
    \bottomrule
  \end{tabular}
\end{table}

The primitives level the tool from its initial pitch of about 9$^\circ$ and re-zero, since the hand's weight would otherwise leave about 0.1\,N$\cdot$m, or 1.3\,cm of center-of-mass error.

\paragraph{Stop and budget.}
The stop of \cref{tab:supp-simconst} binds the harness, the scripts, and the program rows.
It acts when the unfiltered reading crosses the limit, and a no-force tool result says only that the motion was blocked.
Scripts share the episode budget, except 500 turns and 1200 calls for the point-by-point probing of the position-only controller on Peg, and learned policies run for the horizon of \cref{tab:supp-demos}.

\subsection{Task cards}
\label{sec:supp-taskcards}

\Cref{fig:supp-taskcards} gives the task cards, which fix what an episode hides, what the model is told, how the final state is scored, and what the expert force controller of \cref{sec:supp-expert} does.
Every render comes from a recorded episode of that task.

\begin{figure*}[t]
  \centering
  \TaskCard{figs/suppl/task_hp_weight}{Weight}{judge a held load}
    {cube mass, 0.15 or 0.60\,kg.}
    {both masses, place the cube in the matching 16\,cm zone, give a verdict.}
    {cube under 6\,cm above the table and within 8\,cm of the true zone center, fingers open over 3\,cm, no gripper contact.}
    {grasp, \tool{lift} 10\,cm, hold 1\,s, heavy above 0.375\,kg, the class midpoint, from the instruction.}\hfill
  \TaskCard{figs/suppl/task_hp_com}{CoM}{judge a held load}
    {center of mass of a 12\,cm, 0.8\,kg bar, centered or 4\,cm along the axis, in theory 0.31\,N$\cdot$m at the wrist.}
    {the mass and both cases, grasp the bar at its geometric center, place it in the matching zone, give a verdict.}
    {as Weight for the bar center, fingers open over 5\,cm.}
    {as Weight with the tool along the bar, offset above 0.12\,N$\cdot$m, under half of 0.31\,N$\cdot$m, derived.}

  \vspace{7pt}
  \TaskCard{figs/suppl/task_hp_wipe}{Wipe}{find a surface}
    {table height, the minimum of zero and a Gaussian with 3\,cm deviation, lowest top 6.7\,cm below the nominal 0.90\,m.}
    {table up to about 6\,cm low, a spot comes off only at 5\,N or more, more than 15\,N fails.}
    {all 12 spots wiped, each at a normal reading of 5\,N or more, none at 15\,N or more.}
    {fit a line through the remaining spots, \tool{slide} along it at 9\,N, reverse, at most 4 passes, from the instruction.}\hfill
  \TaskCard{figs/suppl/task_hp_stack}{Stack}{find a surface}
    {green cube height, reported with an error uniform within $\pm$1\,cm.}
    {top known to about $\pm$1\,cm, place the red cube gently.}
    {robosuite check, red cube center over 4\,cm above the table, not grasped, touching the green cube.}
    {grasp, \tool{lift}, go 10.5\,cm above the reported cube, descend until contact at 2.5\,N, release, a prior.}

  \vspace{7pt}
  \TaskCard{figs/suppl/task_hp_peg}{Peg}{find a hole}
    {hole position, reported with 15\,mm Gaussian error per axis, square peg clearing the round wall by 4.5\,mm.}
    {clearance about 4\,mm, position off by up to about 3\,cm, feel for the hole, insert at least 2\,cm.}
    {peg center within 7.5\,mm of the hole center, bottom over 2\,cm below the reported hole height, release not checked.}
    {\tool{search} a 5\,cm spiral of 4\,mm pitch at 4\,N, \tool{insert} 2\,cm with a limit of 12 to 15\,N, two retries at higher force, a prior.}\hfill
  \TaskCard{figs/suppl/task_hp_latch}{Latch}{operate a mechanism}
    {unlock angle uniform in 15 to 35$^\circ$, end stop 8$^\circ$ beyond, detent torque rising to 0.45\,N$\cdot$m over the last 8$^\circ$, random hinge damping and friction.}
    {the range, the end stop, the 15\,N line, press the lever until it releases and keep it pressed while pulling, open at least 17$^\circ$.}
    {hinge above 0.3\,rad, no damage.}
    {\tool{turn} the lever until resistance drops, \tool{pull} 3\,cm, turn it back, pull 12\,cm at 8\,N, at most 3 times, lever limit 45$^\circ$, from the instruction and derived.}

  \vspace{7pt}
  \TaskCard{figs/suppl/task_hp_lock}{Lock}{operate a mechanism}
    {locked, or stiff and freed at a handle force uniform in 6 to 11\,N, moving like a locked door until then.}
    {handle already held, both classes, the 6 to 11\,N, the 15\,N line, open a stiff door at least 17$^\circ$ and answer stiff, leave a locked one and answer locked.}
    {no damage, door above 17$^\circ$ if stiff or below 2$^\circ$ if locked.}
    {\tool{pull} 15\,cm along the tangent at 12.5\,N, above 6 to 11\,N and below 15\,N, at most 5 times, stiff if the door moved over 1$^\circ$, from the instruction.}\hfill
  \TaskNote{Shared by all tasks}{Every check is applied in code to the final state.
    On Weight, CoM, and Lock the parity of the seed sets the class, and the verdict also counts as a decision.
    On Latch and Lock, damage is any wrist reading above 15\,N while the gripper is commanded closed, counted on Lock from 1\,s after closing.
    An expert constant comes from the instruction, is derived from it, or is a prior chosen knowing the task and its hidden distribution.}
  \caption{The seven tasks of \bench, each rendered from a recorded episode in MuJoCo. Hidden is what the seed draws and the model never sees, Told summarizes the instruction, Success is the check on the final state, and Expert is the script of the expert force controller with the constant that changes its outcome.}
  \label{fig:supp-taskcards}
\end{figure*}

\paragraph{Scoring on Lock.}
Table 5 of the main paper uses the soft arm with kp 150, and the environment grasps the handle before every episode.
Script rows also need a correct verdict, and harness rows do not but give 20/20 correct verdicts in both conditions.
Learned policies give none, and doing nothing passes every locked door, so 10 of the 11 Diffusion Policy successes are locked doors.

\subsection{Expert and position-only controllers}
\label{sec:supp-expert}

The expert force controller runs one script per task, given on the cards of \cref{fig:supp-taskcards}, and shows how far hand-written control gets with the primitives, noise, and stop of the harness.
Its descents stop at a literature default of 3\,N.
It was written knowing the hidden distributions, so it is a human expert's upper bound rather than a task-free competitor.
Most constants follow from the instruction.
Tuning used seeds disjoint from the test seeds.

\paragraph{Position-only and oracle scripts.}
On the first four tasks the position-only controller runs the expert scripts on the position interface, without \tool{lift} and the contact primitives.
On Peg it probes along a spiral and reads the tool height for a drop, a position-only spiral heuristic.
On Latch a stepped probe adds 2$^\circ$ of lever and tries a 2\,mm pull until the door moves, and an alternating probe alternates a 0.5$^\circ$ hinge turn with 3$^\circ$ of lever.
On Lock a stepped probe moves 1 to 5\,mm along the tangent and calls the door stiff past 0.5$^\circ$, or 0.1$^\circ$.
The oracle position script knows the unlock angle on Latch and the class on Lock.

\subsection{Evidence that the tasks depend on force}
\label{sec:supp-forcedep}

\begin{figure}[t]
  \centering
  \includegraphics[width=\linewidth]{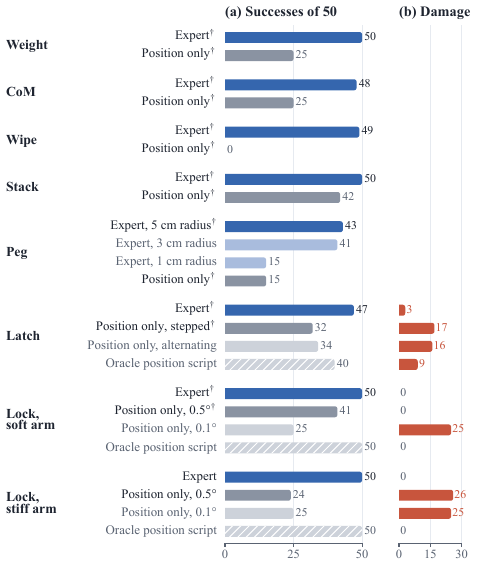}
  \caption{Scripts without a model on the 50 test seeds 9100 to 9149. (a)~Successes and (b)~damaged handles out of 50. Daggers mark the rows of Table 5 of the main paper, lighter bars the variants, and hatched bars the oracle position script. On Lock a success also needs a correct decision.}
  \label{fig:supp-forcedep}
\end{figure}

A task needs force only if kinematics leave the outcome ambiguous, as on Lock, where a still door may be locked or need more force.
\Cref{fig:supp-forcedep} tests this with scripts on all 50 test seeds.
On Weight and CoM the position-only controller gives one fixed answer, right on exactly half of the balanced seeds, while the expert answers all 50 on both and loses two CoM episodes on the final-state check.
On Wipe the position-only controller never reaches the pressure band, and on Peg it runs out of time in 31 episodes.
Stack depends on force least, as its check has no force term.

\paragraph{Knowing the answer is not enough.}
On Latch the oracle position script scores 40/50 against 47/50 for the expert, and 8 of its 9 damaged handles have unlock angles of 29.9$^\circ$ or more, where position targets overrun the 15\,N line.
On Lock with the stiff arm, the 0.5$^\circ$ rule judges all 50 doors correctly yet damages all 25 locked handles and one stiff one, since a 1\,mm probe already presses about 25\,N.
Position targets put no bound on force.

\paragraph{Choice of the position-only row.}
For Latch and Lock, Table 5 of the main paper takes the variant that did best on check seeds disjoint from the test seeds.

\paragraph{What the no-force control can still read.}
We left kinematic cues in place.
A heavy cube lowers the hand by 2 to 3\,mm on Weight, and on Wipe \tool{scene} lists only the remaining spots, a one-bit force reading.
On Lock all 10 locked doors fail because its probe presses a locked handle to about 25\,N unseen.
On CoM we see no leak, as it answers offset in all 20 episodes.

\subsection{Seeds and protocol}
\label{sec:supp-protocol}

Seeds pair the rows.
Table 5 of the main paper uses the test seeds 9100 to 9119, the scripts and both Diffusion Policy rows also run 9100 to 9149, and Code-as-Policies and the other models of \cref{sec:supp-results} run 9100 to 9104.
Demonstrations use seeds 20000 to 20209, up to 20239 on Peg.
Development used 9000 to 9019, and checks, tuning, and earlier model runs used blocks up to 7605, never a test seed.

A seed fixes the hidden variable and the layout, and the logged hidden variables match across rows.

A cell counts successes over 20 seeds by the checks of \cref{fig:supp-taskcards}, after \tool{done} or the budget, or for learned policies after the full horizon with their last checkpoint.
No test seed was dropped.
Intervals are Wilson 95\% intervals and tests are exact McNemar tests on paired seeds.

\section{Extended results}
\label{sec:supp-results}

\begin{figure*}[t]
  \centering
  \includegraphics[width=\textwidth]{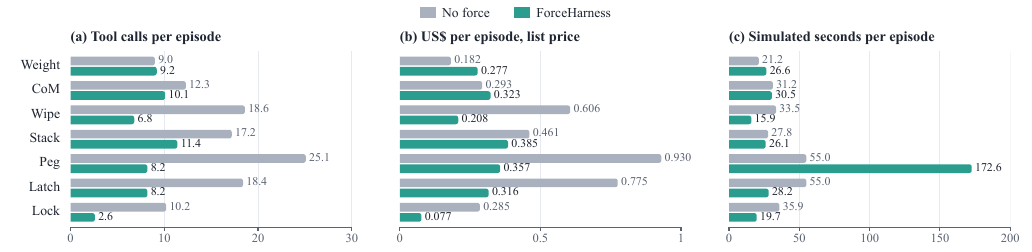}
  \caption{Effort per episode of GPT-6 Astra on each task without force and with \paper, the two harness rows of Table~5 of the main paper. Each bar is a mean over 20 episodes.}
  \label{fig:supp-ours}
\end{figure*}

\begin{figure*}[t]
  \centering
  \includegraphics[width=\textwidth]{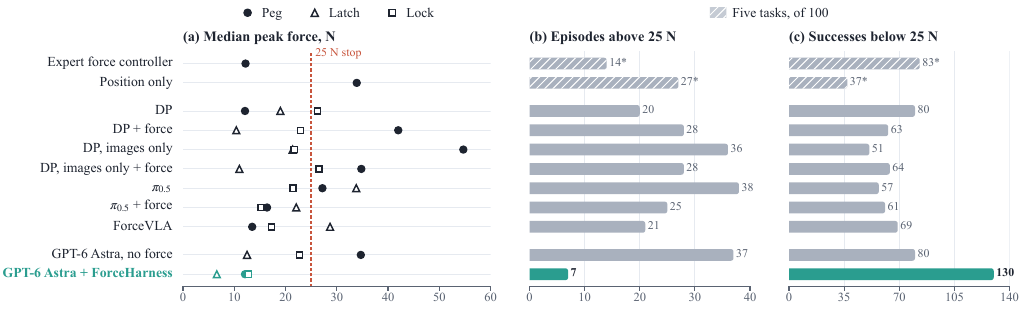}
  \caption{Peak wrist force on \bench. (a)~Median peak per episode on the three tasks where the rows differ, with the 25\,N stop of \paper. (b)~Episodes whose peak passes 25\,N and (c)~successes that stay below it, out of 140. The expert and position-only checks log handle force on Latch and Lock, so their hatched bars cover the other five tasks, 100 episodes.}
  \label{fig:supp-peak}
\end{figure*}

This section expands Tables 5 and 6 of the main paper on \bench, with test seeds 9100 to 9119 unless stated otherwise.
Costs are token counts times list prices without cache discounts.

\subsection{Per-task detail of the two harness rows}
\label{sec:supp-ours}

\Cref{fig:supp-ours} shows the effort behind the two GPT-6 Astra rows of Table~5 of the main paper.
Force cuts both model calls and cost on every task except Weight and CoM.
On Peg the search keeps running, so episodes take longer and three of them ran out of the 300\,s budget.
Without force Astra judges 15 of 20 Weight and 10 of 20 CoM episodes correctly against all 20 with force, and it judges all 20 Lock doors correctly in both rows, so its Lock failures without force are not wrong judgments.

\begin{figure}[t]
  \centering
  \includegraphics[width=\linewidth]{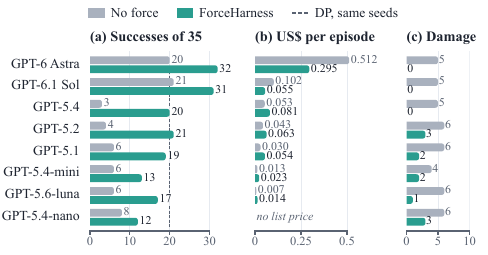}
  \caption{Eight models in the two harness rows on seeds 9100 to 9104, five episodes per task. (a)~Successes out of 35, with Diffusion Policy on the same seeds as the dashed line. (b)~Cost per episode, where GPT-5.4-nano has no listed price. (c)~Damaged handles out of the 10 Latch and Lock episodes.}
  \label{fig:supp-models}
\end{figure}

\begin{figure*}[t]
  \centering
  \includegraphics[width=\textwidth]{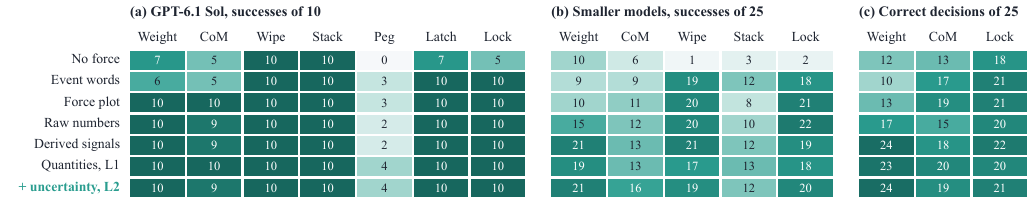}
  \caption{Form of force per task, the rows of Table~6 of the main paper. Each cell counts successes or correct decisions, and its shade gives the rate. Sol runs 10 episodes per task. The five smaller GPT-5.x models run five each, so 25 per task, and were not run on Peg and Latch.}
  \label{fig:supp-form}
\end{figure*}

\begin{figure*}[t]
  \centering
  \includegraphics[width=\textwidth]{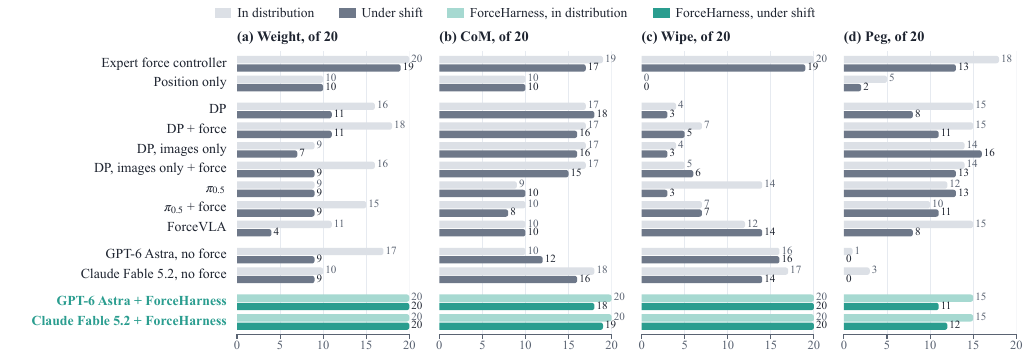}
  \caption{Successes per task in distribution and under shift on seeds 9100 to 9119, out of 20. Their totals are the Shift column of Table~5 of the main paper.}
  \label{fig:supp-shift}
\end{figure*}

\subsection{Peak force}
\label{sec:supp-peak}

\Cref{fig:supp-peak} shows how hard each row presses, beyond the damaged handles counted in Table~5 of the main paper.
On the sorting, wiping, and stacking tasks every median peak lies between 2.2 and 19.1\,N, so the figure gives only Peg, Latch, and Lock.
\paper has the fewest episodes above 25\,N, 7 of 140, and its three on Lock, which the no-force row shares, come from the scripted pre-grasp, since every per-call peak there stays below 13\,N.

\subsection{Learned baselines}
\label{sec:supp-dp50}
\label{sec:supp-dpinputs}
\label{sec:supp-vla}

Further runs test the learned rows of Table~5 of the main paper over more seeds and with fewer demonstrations.
Over the 50 seeds 9100 to 9149, Diffusion Policy succeeds in 203/350 episodes without force and 204/350 with it, against 337/350 for the scripts that recorded its demonstrations.
With force its median peak on Peg rises from 12.4 to 40.0\,N, with 28 of 50 episodes past 25\,N.
Doing nothing succeeds on a locked door, so most of its Lock score is free, while stiff doors succeed in only 3 and 4 of 25 without and with force.
Going from 200 to 50 demonstrations costs 21 of 140 episodes with and without force.
Prompt Fusion makes $\pi_{0.5}$ press more gently but leaves its total at 69 and 67.

\subsection{Other models}
\label{sec:supp-models}

\Cref{fig:supp-models} repeats the two harness rows on eight models with five seeds per task, and all eight improve with force.
Sol nearly matches Astra at about a fifth of the cost, and GPT-5.4, GPT-5.2, and GPT-5.1 reach 19 to 21, close to Diffusion Policy at 20 and 18 without and with force on the same seeds.
With force, damaged handles fall for every model, while the cost per episode falls for Astra and Sol and rises for the GPT-5.x models.
Only Astra and Sol solve Latch, and Peg stays the hardest task for every model.

\begin{table}[t]
  \caption{Code-as-Policies per task, as successes out of 5 on seeds 9100 to 9104, against the harness rows and the expert.}
  \label{tab:supp-cap}
  \centering
  \footnotesize
  \setlength{\tabcolsep}{4.5pt}
  \begin{tabular}{@{}lccccc@{}}
    \toprule
    & \multicolumn{2}{c}{One program} & \multicolumn{2}{c}{Step by step} & \\
    \cmidrule(lr){2-3}\cmidrule(lr){4-5}
    Task & None & Force & None & Force & Expert \\
    \midrule
    Weight & 0 & 5 & 3 & 5 & 5 \\
    CoM & 0 & 4 & 3 & 5 & 4 \\
    Wipe & 3 & 5 & 4 & 5 & 5 \\
    Stack & 0 & 3 & 5 & 5 & 5 \\
    Peg & 0 & 0 & 0 & 2 & 4 \\
    Latch & 2 & 5 & 3 & 5 & 4 \\
    Lock & 3 & 5 & 2 & 5 & 5 \\
    \midrule
    Total & 8 & 27 & 20 & 32 & 32 \\
    \bottomrule
  \end{tabular}
\end{table}

\subsection{Code-as-Policies per task}
\label{sec:supp-cap}

\Cref{tab:supp-cap} splits the Code-as-Policies result of Figure~4c of the main paper by task.
Without force, a program written in one call falls behind mostly on Stack and the two sorting tasks, where it cannot change course after a tool result.
With force it reads our quantities, so its 27/35 measures such a program over our force interface, as \cref{sec:supp-baselines} explains.
None of the 70 program episodes had a program error or damaged a handle.

\subsection{Form of force}
\label{sec:supp-form}

\Cref{fig:supp-form,tab:supp-formpairs} expand Table~6 of the main paper per task and on paired episodes, with the same force primitives in every row except no force.
Sol runs on seeds 9100 to 9109 and the GPT-5.x models of \cref{fig:supp-models} except GPT-5.4-nano on 9100 to 9104, reusing their no-force and L2 runs.
For Sol, Peg is the only task that stays hard.
For the smaller models, the gain of computed quantities in decisions is clearest on Weight, from 12 of 25 without force to 17 with raw numbers and 24 with L2.
Since derived signals do as well as L1 and L2, we credit computing the quantity rather than its format.
Raw numbers here are tared and show the last eight samples.

\begin{table}[t]
  \caption{Paired comparisons with the same model, task, and seed. Each cell gives the episodes where only the first form succeeds, then those where only the second does, and the exact McNemar $p$.}
  \label{tab:supp-formpairs}
  \centering
  \footnotesize
  \setlength{\tabcolsep}{2.5pt}
  \begin{tabular}{@{}lccc@{}}
    \toprule
    & Sol & Smaller & Smaller \\
    Comparison & success & success & decisions \\
    \midrule
    L2, raw & 4, 2, 0.69 & 23, 14, 0.19 & 18, 6, 0.023 \\
    L2, events & 11, 2, 0.023 & 34, 13, 0.003 & 24, 8, 0.007 \\
    L2, plot & 3, 3, 1 & 28, 10, 0.005 & 19, 8, 0.052 \\
    L2, signals & 3, 1, 0.63 & 18, 16, 0.86 & 9, 9, 1 \\
    L2, L1 & 1, 2, 1 & 20, 12, 0.22 & 8, 7, 1 \\
    L2, no force & 20, 1, $2\times 10^{-5}$ & 71, 5, $5\times 10^{-16}$ & 29, 8, 0.0008 \\
    Raw, no force & 18, 1, $8\times 10^{-5}$ & 63, 6, $4\times 10^{-13}$ & 23, 14, 0.19 \\
    Signals, raw & -- & -- & 16, 4, 0.012 \\
    L1, raw & -- & -- & 18, 7, 0.043 \\
    \bottomrule
  \end{tabular}
\end{table}

\begin{figure*}[!t]
  \centering
  \includegraphics[width=0.9\linewidth]{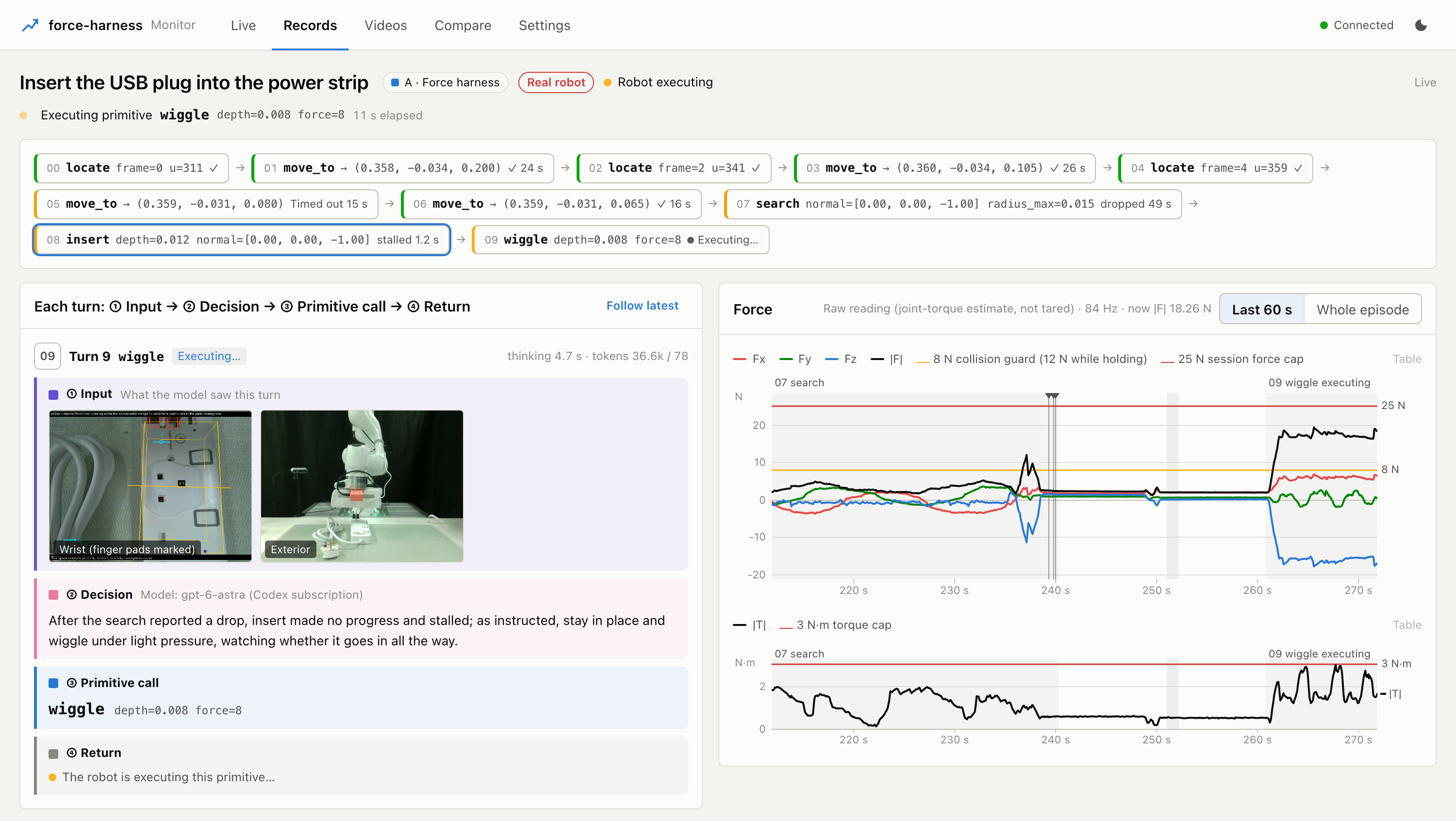}
  \fixedfigure{E1}
  \caption{The web monitor while the robot runs \tool{wiggle} during a USB-plug insertion.
  The strip at the top lists the calls so far, the left column follows the current turn, and the right column plots wrist force and torque over the last 60\,s.
  The 8\,N line guards free motion, so a force primitive may press above it, up to the 25\,N cap.
  The page replays the recorded episode of \cref{fig:supp-monitor-episode} in real time.}
  \label{fig:supp-monitor-live}
  \addtocounter{figure}{-1}
\end{figure*}

\begin{figure*}[!t]
  \centering
  \begin{minipage}[t]{0.36\linewidth}\vspace{0pt}
    \centering
    \includegraphics[height=7.8cm]{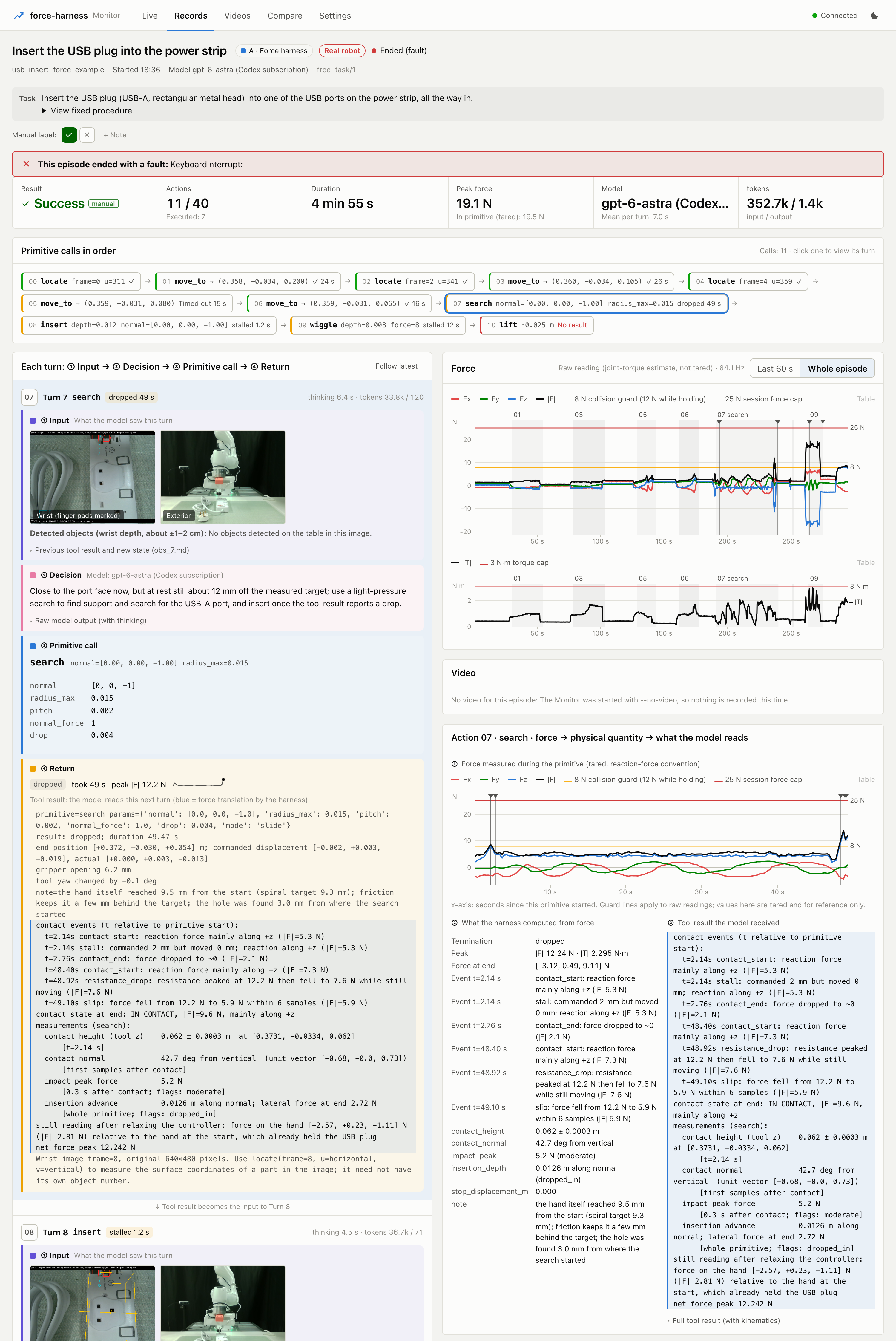}
    \fixedfigure{E2}
    \captionof{figure}{The page of the same episode after it ended, with the \tool{search} turn selected.
    The operator interrupted the episode and labeled it a success by hand, so the page shows both a fault and a manual label, and the video panel is empty because the replay has no video.}
    \label{fig:supp-monitor-episode}
  \end{minipage}\hfill
  \begin{minipage}[t]{0.61\linewidth}\vspace{0pt}
    \centering
    \includegraphics[height=5.7cm]{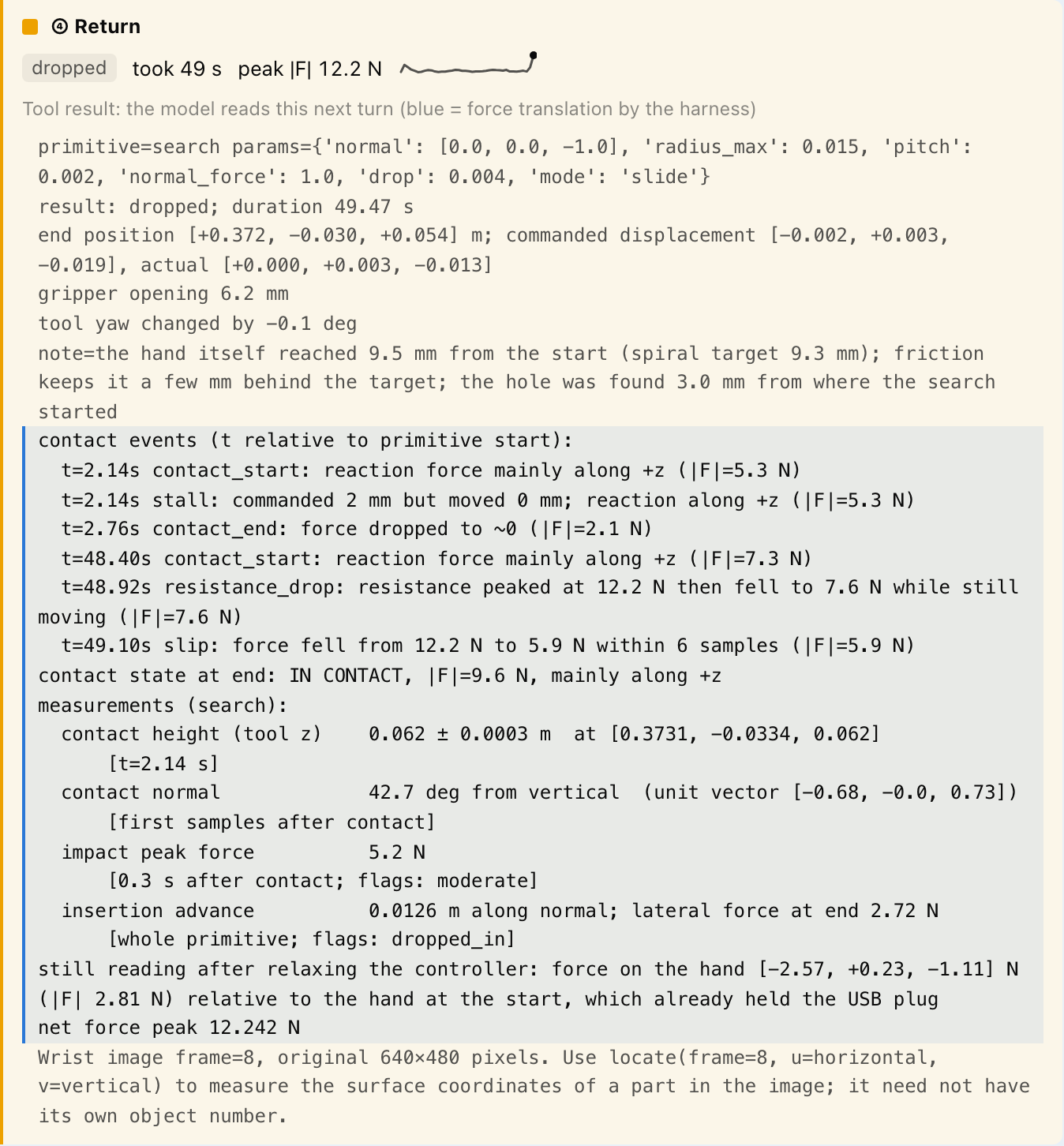}\hspace{8pt}%
    \includegraphics[height=5.7cm]{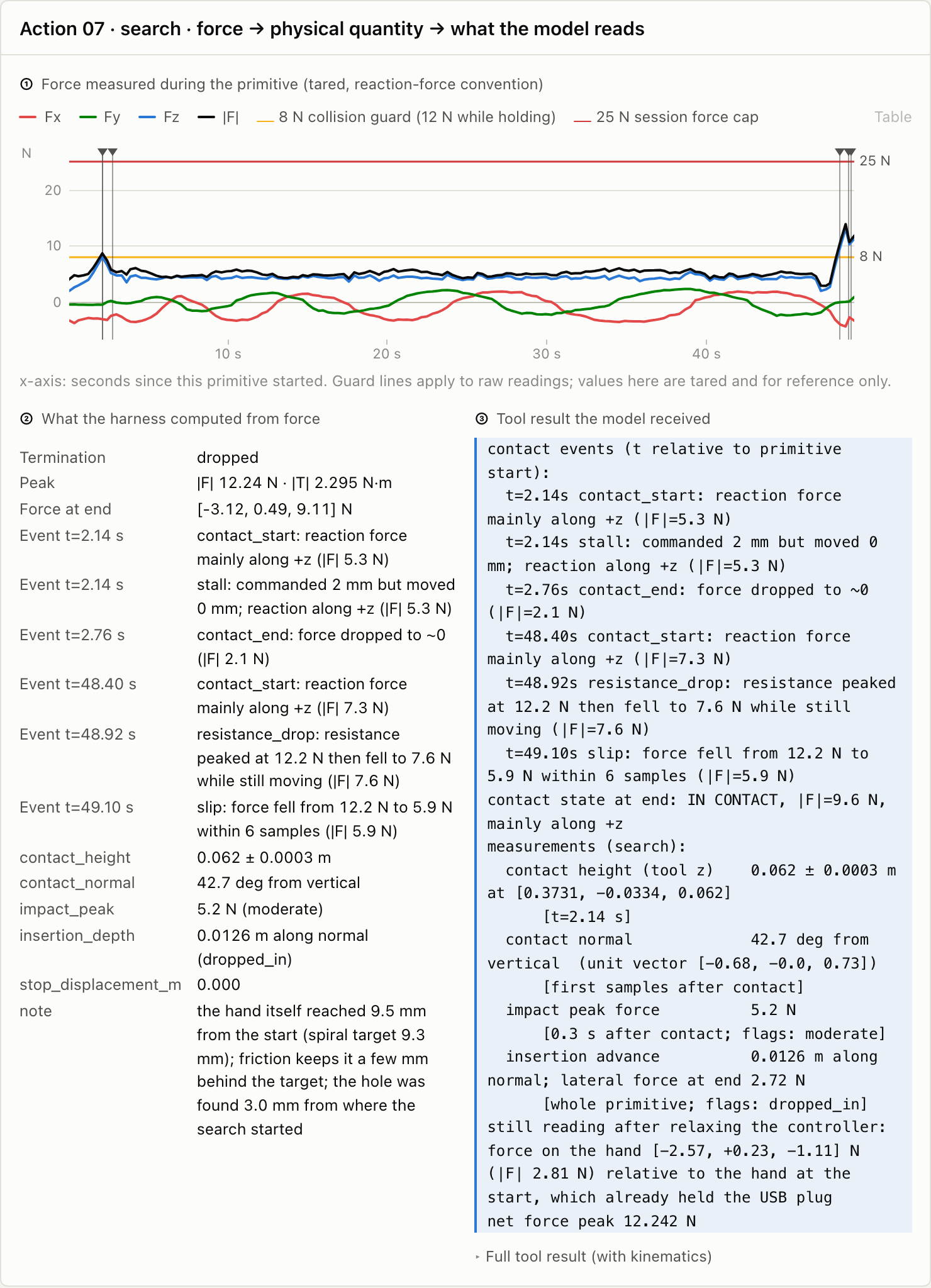}
    \fixedfigure{E3}
    \captionof{figure}{The \tool{search} turn of \cref{fig:supp-monitor-episode}, enlarged.
    On the left is the returned tool result, whose shaded block with a blue edge is what the physics translation layer added.
    On the right are the action details, which set the recorded force beside what the harness computed from it and the tool result that the model received.}
    \label{fig:supp-monitor-detail}
  \end{minipage}
  \addtocounter{figure}{-2}
\end{figure*}

\subsection{Distribution shift per task}
\label{sec:supp-shift}

The shifted variants change only the hidden variables and the numbers in the instruction.
Weight uses masses of 0.30 and 0.45\,kg instead of 0.15 and 0.60\,kg, CoM an offset of 2.5\,cm instead of 4\,cm, Wipe a table up to 3\,cm lower than before, and Peg a hole noise of 25\,mm instead of 15\,mm, while object names, the scene format, and the success checks stay the same.
No policy is retrained and the expert keeps its constants.
\Cref{fig:supp-shift} expands the Shift column of Table~5 of the main paper per task.

Paired by seed, \paper with GPT-6 Astra has 8 episodes that succeed only in distribution and 2 only under shift, and its 69 against the expert's 68 is a tie.
With Claude Fable 5.2 the counts are 5 and 1, while Fable without force loses on all four tasks.
Over 50 seeds, the expert goes from 190 to 175 of 200 and the position-only script from 65 to 59, while Diffusion Policy goes from 128 to 107 without force and from 151 to 117 with it.
On Wipe, $\pi_{0.5}$ falls from 14 to 3 while force-reading rows hold.

Under shift \paper misjudged two CoM episodes, reading offsets near 1.1\,cm when the truth was 0 or 2.5\,cm.
On Peg, most of its failures begin with a first search that settles away from the true hole under the larger position noise.
Only the frozen models read the new ranges from their instruction.
Without force Astra loses most on Weight, while the image-only Diffusion Policy loses least because it never reads the noisier hole position and its wrist camera still sees the hole.

\section{Web monitor}
\label{sec:supp-monitor}

The web monitor of Sec.~5.1 of the main paper lets a reader check any tool result against the force behind it.
\Cref{fig:supp-monitor-live} shows it while a primitive runs, and \cref{fig:supp-monitor-episode,fig:supp-monitor-detail} show the same episode after it ended.

The force panel plots the three components of $F$ and $|F|$, live or over a whole episode, and marks each primitive with a band, each contact event with a triangle, and each guard stop of \cref{sec:supp-prim-constants} with a line.
A second panel plots torque.
The curves are raw and not tared, so a bias of about 1 to 2\,N is normal.
Each turn is a card from the model's input and note to the returned tool result, and selecting it opens the action details, which is how we checked that each reported quantity matches the recorded force.

The monitor only reads run files and relays messages between the harness and the model, and all limits and checks stay in the harness.

\section{What the model sees}
\label{sec:supp-prompts}

This section shows what a frozen model reads on \bench under each form of force of Table 6 of the main paper and under the no-force control.
Excerpts are copied from stored transcripts or source files, and three dots mark cuts.

\subsection{System prompt}
\label{sec:supp-prompts-system}

All conditions share one system prompt, the template of \cref{fig:supp-prompts-system} with five placeholders in braces.

\begin{figure}[ht]
{\scriptsize
\begin{verbatim}
You control a Franka Panda arm in a tabletop
  manipulation simulation through a small set of motion
  primitives (tools).
You are not in the control loop: each primitive runs its
  own closed-loop controller at 20 Hz and returns when
  it finishes,
when it is blocked, or when a safety limit trips. You
  decide what to do between primitives.
...
Action interface for this episode: {interface}.
  Available tools: {tool_names}.
Sensing you receive: {force_note} {vision_note}
...
2. Prefer small, deliberate motions near objects. Any
  motion whose wrist force exceeds {f_cap:g} N is
  aborted by the safety layer.
3. When something is uncertain and cannot be seen, you
  may probe it physically (touch, lift, push) if your
  tools allow it.
4. Be honest in done: report success=true only if you
  believe the task is actually complete; give a verdict
  when the task asks for one.
...
\end{verbatim}
}
  \caption{The system prompt shared by all conditions, verbatim without its frames and units, its wrist-sensor line, and rules 1 and 5.}
  \label{fig:supp-prompts-system}
\end{figure}

Here \code{force\_note} is the sentence of \cref{tab:supp-prompts-notes} for the form of force, \code{interface} is \code{force} or \code{pose}, and \code{tool\_names} lists the tools of the task and interface from \cref{tab:supp-prompts-tools}.
\code{f\_cap} is 25, and in every MuJoCo row of the main paper \code{vision\_note} reads ``You receive NO camera images; rely on the scene description and the signals you are given.''
The first user message adds the instruction, the scene as JSON, and the robot state, and a rejected call leaves the robot still and returns a \code{tool error:} line.
The prompts of \paper and the no-force control thus differ only in the interface, the tool list, and the force sentence, and both mention the wrist sensor and the same 25\,N limit.
The FR3 prompt adds that force is estimated from joint torques and that weights should be compared within an episode.

\begin{table}[tb]
  \caption{The sentence that each form of force adds to the system prompt, verbatim. Every form after no force opens with ``After each primitive you receive'', cut at the three dots, and L2 appends its sentence to that of L1.}
  \label{tab:supp-prompts-notes}
  \centering
  \footnotesize
  \setlength{\tabcolsep}{3pt}
  \let\rr\raggedright
  \begin{tabular}{@{}p{0.22\linewidth}p{0.75\linewidth}@{}}
    \toprule
    Form & Sentence \tabularnewline
    \midrule
    \rr No force & \rr You receive NO force or torque information; only motion results. \tabularnewline
    \rr Event words & \rr ... a list of detected contact EVENTS (contact start/end, safety cap, stall, slip, resistance drop) and whether the tool is in contact at the end. \tabularnewline
    \rr Force plot & \rr ... a PLOT (image) of the wrist force and torque versus time. \tabularnewline
    \rr Raw numbers & \rr ... the raw wrist force/torque numbers (current value and recent samples). \tabularnewline
    \rr Derived signals & \rr ... detected contact EVENTS plus derived signals (peak force, contact direction, payload estimate when lifting, mean normal force when pushing/sliding, torque peak when rotating). \tabularnewline
    \rr Quantities, L1 & \rr ... detected contact EVENTS plus physical MEASUREMENTS computed from the wrist force/torque (payload mass, centre-of-mass offset of the held load, contact height/point, ...), each with units. \tabularnewline
    \rr + uncertainty, L2 & \rr Each measurement carries an uncertainty ($\pm$) and quality flags. \tabularnewline
    \bottomrule
  \end{tabular}
\end{table}

\begin{table*}[t]
  \caption{Tool schemas on \bench, as the model receives them. Descriptions are verbatim and cut at the three dots, a question mark marks an optional parameter, and G and H stand for the interface sentences in the last rows.}
  \label{tab:supp-prompts-tools}
  \centering
  \footnotesize
  \setlength{\tabcolsep}{3pt}
  \let\rr\raggedright
  \begin{tabular}{@{}p{0.135\linewidth}p{0.335\linewidth}p{0.505\linewidth}@{}}
    \toprule
    Tool & Parameters & Description \tabularnewline
    \midrule
    \multicolumn{3}{@{}l}{\textit{Both interfaces, \code{pose} and \code{force}}} \tabularnewline
    \rr \tool{scene} & \rr none & \rr Get the current scene description ... and the robot state. \tabularnewline
    \rr \tool{move\_to} & \rr \code{x, y, z, speed?} & \rr Move the tool centre point to an absolute world position ... G \tabularnewline
    \rr \tool{move\_by} & \rr \code{dx, dy, dz, speed?} & \rr Move the tool centre point by a relative offset ... G \tabularnewline
    \rr \tool{rotate\_z} & \rr \code{angle\_deg, torque\_cap?, stop\_on\_resistance\_drop?} & \rr Rotate the tool about the vertical axis by angle\_deg ... H \tabularnewline
    \rr \tool{grasp} & \rr none & \rr Close the gripper. ... \tabularnewline
    \rr \tool{release} & \rr none & \rr Open the gripper. \tabularnewline
    \rr \tool{hold} & \rr \code{seconds?} & \rr Stay still for seconds seconds (measures the wrist signal while holding). \tabularnewline
    \rr \tool{point\_tool} & \rr \code{direction} & \rr Re-orient the tool so the fingers point along direction ... G \tabularnewline
    \rr \tool{turn} & \rr \code{axis, pivot, angle\_deg, rate\_deg\_s?, torque\_cap?, stop\_on\_resistance\_drop?} & \rr Rotate the hand rigidly by angle\_deg about the axis that passes through pivot ... H \tabularnewline
    \rr \tool{done} & \rr \code{success, verdict?, note?} & \rr Declare the episode finished. ... \tabularnewline
    \midrule
    \multicolumn{3}{@{}l}{\textit{\code{force} interface only}} \tabularnewline
    \rr \tool{move\_until\_\allowbreak contact} & \rr \code{direction, max\_dist, f\_threshold?} & \rr Move slowly along a direction vector until the wrist senses contact ... \tabularnewline
    \rr \tool{push} & \rr \code{direction, target\_force, duration?} & \rr Press along a direction and regulate the contact force ... \tabularnewline
    \rr \tool{slide} & \rr \code{direction, distance, normal\_force?, normal?} & \rr Slide along a surface: keep normal\_force newtons against the surface ... \tabularnewline
    \rr \tool{search} & \rr \code{normal?, radius\_max?, pitch?, normal\_force?, drop?, mode?, hop?, speed?} & \rr Insertion search around the current position ... Stops with result 'dropped' as soon as the tool advances more than drop metres along the normal ... \tabularnewline
    \rr \tool{insert} & \rr \code{depth?, normal?, max\_force?} & \rr Compliant insertion along normal (default downward): ... \tabularnewline
    \rr \tool{lift} & \rr \code{height?, hold\_time?} & \rr Raise the tool vertically by height metres, then hold still ... \tabularnewline
    \rr \tool{probe} & \rr \code{direction?, amplitude?, f\_threshold?} & \rr Tap: move along direction until contact ... and return to the start. ... \tabularnewline
    \rr \tool{pull} & \rr \code{direction, force?, distance?, speed?} & \rr Pull something you are holding that is attached to a mechanism ... \tabularnewline
    \midrule
    \multicolumn{3}{@{}l}{\textit{Interface sentences}} \tabularnewline
    \rr G, \code{force} & \multicolumn{2}{@{}p{0.85\linewidth}@{}}{The motion STOPS EARLY if the tool makes contact with anything (contact guard) and reports it.} \tabularnewline
    \rr G, \code{pose} & \multicolumn{2}{@{}p{0.85\linewidth}@{}}{The motion completes the full commanded displacement unless a safety force limit is hit.} \tabularnewline
    \rr H, \code{force} & \multicolumn{2}{@{}p{0.85\linewidth}@{}}{Optionally stop when torque exceeds torque\_cap or when resistance peaks and then drops.} \tabularnewline
    \rr H, \code{pose} & \multicolumn{2}{@{}p{0.85\linewidth}@{}}{Rotation completes fully unless a safety torque limit is hit.} \tabularnewline
    \bottomrule
  \end{tabular}
\end{table*}

\begin{figure*}[t]
\begin{minipage}[t]{0.49\linewidth}
{\scriptsize
\begin{verbatim}
== Event words, full result
primitive=lift params={'height': 0.12, 'hold_time': 2}
result: completed
end position [-0.015, -0.004, +0.945] m; commanded
  displacement [+0.000, +0.000, +0.127], actual [-0.003,
  -0.000, +0.123]
tool yaw changed by -0.1 deg
note=a torque about a horizontal axis while holding
  still means the load's centre of mass is offset from
  the grasp
contact events: none (no contact detected during this
  primitive)
contact state at end: IN CONTACT
robot state: tool at [-0.015, -0.004, +0.945] m, gripper
  closed
== No force, a move_by since this interface has no lift
primitive=move_by params={'dx': 0, 'dy': 0, 'dz': 0.12}
result: completed
end position [-0.015, -0.004, +0.938] m; commanded
  displacement [+0.000, +0.000, +0.120], actual [-0.003,
  -0.000, +0.117]
...
== Force plot
...
wrist force/torque during this primitive is shown in the
  attached plot (world frame, N / Nm, bias removed;
  dotted vertical lines mark detected events).
<image attached>
...
\end{verbatim}
}
\end{minipage}\hfill
\begin{minipage}[t]{0.49\linewidth}
{\scriptsize
\begin{verbatim}
== Raw numbers
...
  current: Fx=-1.32 Fy=-0.33 Fz=-7.83 Tx=-0.09 Ty=-0.25
    Tz=-0.00
  last 8 samples (t_rel, Fx, Fy, Fz, Tx, Ty, Tz):
    +8.10, -1.06, +1.01, -7.82, -0.11, -0.27, -0.00
    ...
== Derived signals
...
peak |F| during primitive: 9.0 N; peak |T|: 0.40 Nm
derived signals: hold_force_z_N=-7.78,
  hold_force_z_std_N=0.52, payload_estimate_kg=0.793,
  hold_torque_xy_Nm=[-0.054, -0.246],
  hold_torque_xy_mag_Nm=0.252, ...
== Quantities, L1
...
measurements (lift):
  payload mass               0.793 kg
  centre-of-mass offset      3.22 cm  toward mostly +x
    (away from the base)
...
== + uncertainty, L2, the default of ForceHarness
...
measurements (lift):
  payload mass               0.791 ± 0.032 kg
      [last 40 samples, 2.0 s]
  centre-of-mass offset      3.32 ± 0.29 cm  toward
    mostly +x (away from the base)
      [last 40 samples, 2.0 s]
...
\end{verbatim}
}
\end{minipage}
  \caption{The same \tool{lift} on CoM at test seed 9100 under each form of force, verbatim, with one episode per form. Lines starting with \code{==} are our labels, three dots outside Event words stand for the shared kinematic, contact, and robot-state lines, and \code{\textless image attached\textgreater} stands for a PNG that plots force and torque against time.}
  \label{fig:supp-prompts-forms}
\end{figure*}

\subsection{Tool schemas}
\label{sec:supp-prompts-tools}

\Cref{tab:supp-prompts-tools} lists the tools of \bench as the model receives them, and the interface changes only the closing sentence of the motion tools and the rotations, marked G and H.
Wipe has no gripper and drops \tool{grasp}, \tool{release}, and \tool{lift}, while \tool{point\_tool} and \tool{pull} appear on Lock and Latch only and \tool{turn} on Latch only.
A third interface, the \code{pose} tools with a stop on contact, serves only the real-robot no-force control of \cref{sec:supp-real}, and ForceBench adds the tools of \cref{sec:supp-forcebench}.

\subsection{Tool results under each form of force}
\label{sec:supp-prompts-results}

\Cref{fig:supp-prompts-forms} shows one \tool{lift} on CoM, where the bar's true offset is 4\,cm, under the seven forms of Table 6 of the main paper, with GPT-6.1 Sol at test seed 9100.
All seven episodes answer OFFSET, which is correct, but without force the model answers OFFSET at every seed from 9100 to 9104, whatever the truth.
Event words print $|F|$ only at an event, so a still hold gets no reading, and on Weight GPT-6.1 Sol with event words then calls 4 of the 5 heavy cubes light.
Derived signals give the torque in N\,m and leave the lever to the model, while L1 and L2 give the offset in centimeters.

\paragraph{Proprioception.}
\label{sec:supp-prompts-proprio}
Kinematics can leak force, so on \bench and ForceBench every form omits the primitive's duration and, while carrying, the gripper opening, which change with the load, but keeps the end position and displacement.
The hand still sags by 2 to 3\,mm under a heavy cube and 1\,mm under a light one, so the no-force control is not blind to force and may score above a truly force-blind model.
We did not hide more, to avoid an artificially ignorant control.
On the FR3, tool results keep the duration and the gripper opening in both conditions.

\subsection{One short episode as a trace}
\label{sec:supp-prompts-trace}

A short Lock episode at test seed 9100 shows both conditions with GPT-6.1 Sol and the same instruction and scene, and the model wrote no text besides its tool calls.

\par\smallskip\noindent\begin{minipage}{\linewidth}
{\scriptsize
\begin{verbatim}
TASK: Your gripper is already holding the door handle.
  This door is either LOCKED or just STIFF (if stiff, it
  takes 6-11 N at the handle to get it moving). ... If
  it is locked, do not force it: call done with verdict
  'locked'. Any force above 15 N on the handle damages
  it and fails the task. ...
\end{verbatim}
}
\end{minipage}
\par\noindent\begin{minipage}{\linewidth}
{\scriptsize
\begin{verbatim}
== ForceHarness
call 1: pull {"direction": [-0.12, 0.993, 0], "force":
  12, "distance": 0.115, "speed": 0.015}
...
result: stalled
...
measurements (pull):
  mean pulling force         8.82 ± 0.84 N; sideways
    force 1.75 N
      [last 80%; flags: mechanism_stopped]
...
call 2: done {"success": true, "verdict": "locked",
  "note": "The door did not rotate under a controlled 12
  N pull, above the stated stiff-door breakaway range
  and below the 15 N damage threshold."}
\end{verbatim}
}
\end{minipage}
\par\noindent\begin{minipage}{\linewidth}
{\scriptsize
\begin{verbatim}
== No-force control: calls 1 to 4 completed
...
call 5: move_to {"x": -0.1402, "y": -0.247, "z": 1.078,
  "speed": 0.001}
...
result: stopped early (motion blocked, safety stop)
...
call 6: done {"success": true, "verdict": "locked",
  "note": "Handle remained immobile under small, slow
  opening-direction probes; stopped probing at
  motion-blocked indication."}
\end{verbatim}
}
\end{minipage}\par\smallskip

Under \paper the hand does not move and the primitive stops itself at a mean of 8.82\,N, and the record gives a 12.58\,N peak handle force, no damage, and success.
Under no force, tool results print no force, events, or contact state, and a stop on force loses its number, so the fifth says only that a safety stop ended the motion, which the prompt ties to 25\,N.
The model reports success with the correct verdict, but the record gives a 25.12\,N peak, damage, and failure.

\section{Baselines}
\label{sec:supp-baselines}

This section describes how the baseline rows of Table 5 of the main paper were built, with the training recipes in \cref{tab:supp-dprecipe} and the demonstrations in \cref{tab:supp-demos}.
We train every learned baseline on \bench ourselves, each with one training seed.

\begin{table*}[t]
  \caption{Training recipes of the learned policies on \bench. All three see two camera views and predict a pose delta of 7 dims plus 6 stiffness values, and each is trained with one seed and evaluated at its last step, with no checkpoint selection.}
  \label{tab:supp-dprecipe}
  \centering
  \footnotesize
  \setlength{\tabcolsep}{4pt}
  \let\rr\raggedright
  \begin{tabular}{@{}lp{0.335\linewidth}p{0.29\linewidth}p{0.265\linewidth}@{}}
    \toprule
    & Diffusion Policy & $\pi_{0.5}$ & ForceVLA \tabularnewline
    \midrule
    Model & \rr ResNet18 per view from scratch, GroupNorm, spatial softmax, 32 keypoints, 1D U-Net 256, 512, 1024 with FiLM, 100 DDPM and 16 DDIM steps & \rr \code{pi05\_base}, all parameters trained & \rr $\pi_{0}$ from \code{pi0\_base}, force token, LIMoE of 4 experts, ForceBench code \tabularnewline
    State & \rr proprioception 11, scene 5 per object & \rr proprioception 11, scene 15, as prompt tokens after quantile normalization & \rr continuous, with the noisy scene \tabularnewline
    Force & \rr 6 dims in the world frame, zeroed at reset, noisy & \rr 6 dims as prompt tokens & \rr force token \tabularnewline
    Action & \rr 6 pose dims on Wipe, 2 frames observed, 16 predicted, 8 executed & \rr 10 predicted, 5 executed, min-max for stiffness, pose deltas widened to $|z|\le 5$ & \rr 13 dims, not the 7-dim absolute pose of ForceBench \tabularnewline
    Data & \rr one model per task, first 200 saved episodes by seed, 185 on Peg, 197 on Latch & \rr one model for 7 tasks, all 1423 episodes, task selected by prompt & \rr as $\pi_{0.5}$ \tabularnewline
    Training & \rr 40\,000 steps, batch 64, AdamW at 1e-4, cosine, 500 warmup steps, EMA & \rr 30\,000 steps, batch 32, seed 42, AdamW, peak 5e-5, 10\,000 warmup steps, EMA 0.999 & \rr 30\,000 steps, global batch 64, optimizer and seed of ForceBench \tabularnewline
    \bottomrule
  \end{tabular}
\end{table*}

\subsection{Demonstrations}
\label{sec:supp-demos}

All learned policies train on the successful episodes of the expert force controller, counted in \cref{tab:supp-demos}, using all 185 on Peg.
Lock uses the soft arm and records after the pre-grasp.
The recorder stores the action that the environment received plus six stiffness values, because the primitives switch stiffness and a policy with fixed stiffness could not learn the expert's contact behavior.
Force is stored noise-free and receives the noise of \cref{sec:supp-simsetup} at conversion.

\begin{table}[!htb]
  \caption{Expert episodes recorded per task and the successful ones saved for training, with lengths in ticks at 20\,Hz. The rollout horizon of a learned policy is 1.5 times the longest demonstration of the task.}
  \label{tab:supp-demos}
  \centering
  \footnotesize
  \setlength{\tabcolsep}{4pt}
  \begin{tabular}{@{}lccccc@{}}
    \toprule
    Task & Attempts & Saved & Frames & Longest & Horizon \\
    \midrule
    Weight & 210 & 210 & 144\,312 & 723 & 1084 \\
    CoM & 210 & 208 & 140\,607 & 709 & 1063 \\
    Wipe & 210 & 207 & 68\,960 & 453 & 679 \\
    Stack & 210 & 206 & 121\,311 & 793 & 1189 \\
    Peg & 240 & 185 & 122\,194 & 2310 & 3465 \\
    Latch & 210 & 197 & 111\,329 & 673 & 1009 \\
    Lock & 210 & 210 & 18\,154 & 254 & 381 \\
    \midrule
    Total & 1500 & 1423 & 726\,867 & & \\
    \bottomrule
  \end{tabular}
\end{table}

\subsection{Training recipes}
\label{sec:supp-dprecipe}
\label{sec:supp-pirecipe}
\label{sec:supp-fvrecipe}

Each learned policy departs from its published recipe as described below, and one evaluation client applies the damage rules of \cref{sec:supp-taskcards} to all of them.

\paragraph{Diffusion Policy.}
Our implementation of the CNN Diffusion Policy~\cite{chi2023dp} departs from its robomimic image setup in using 16-step DDIM sampling, stiffness in the action, and images of 128 resized to 96 and cropped to 88 rather than 84 and 76.
Force is concatenated with the low-dimensional state, matching the concatenation baseline in FoAR~\cite{he2025foar} and ForceVLA~\cite{yu2025forcevla}.
The images-only rows drop the scene vector but keep proprioception, as ForceBench does, and equal the full input on Wipe, whose scene vector is empty.
The scene vector is computed once and holds no door or lever angle, while a frozen model can call \tool{scene} at any time.

\begin{table*}[t]
  \fixedtable{H1}
  \caption{The eight tasks. $C$ is the sampled angle of the resistance peak, the detent, $d$ the hidden sign of the easier direction, and $x_t$ the opening at 2.35\,N of designed resistance. The third column is the public evaluator at commit \code{31acf73}, which produced our rows, and the last column what Appendix A of the ForceBench paper requires beyond it.}
  \label{tab:supp-fb-tasks}
  \label{tab:supp-fb-info}
  \addtocounter{table}{-1}
  \centering
  \footnotesize
  \setlength{\tabcolsep}{4pt}
  \let\rr\raggedright
  \begin{tabular}{@{}lp{0.17\linewidth}p{0.43\linewidth}p{0.235\linewidth}@{}}
    \toprule
    Task & Hidden variable & Success in the evaluator we ran & Also required in the paper \tabularnewline
    \midrule
    Weight Sort & \rr mass class & \rr correct container, release, settling & -- \tabularnewline
    CoM Sort & \rr CoM centered or 2\,cm off & \rr correct destination, release, settling & -- \tabularnewline
    Faucet Turn & \rr $C$ from 20$^\circ$ to 30$^\circ$ & \rr angle past $C+0.95^\circ$, water visible & \rr final angle in $[C, C+2.1^\circ]$, release, clearance, rest \tabularnewline
    Faucet Direction & \rr $C$ and $d$ & \rr final $d\theta_f \ge C$, water visible, release, clearance, rest & -- \tabularnewline
    Stove Turn & \rr $C$ from 60$^\circ$ to 90$^\circ$ & \rr crossing at $C+0.5^\circ$, final angle within 3$^\circ$, flame visible & \rr release, clearance \tabularnewline
    Stove Press & \rr $C$ and $d$ & \rr $d\theta \ge C+0.5^\circ$ while pressed & -- \tabularnewline
    Drawer Pull & \rr resistance profile & \rr opening within 2.5\,mm of $x_t$ for 15 slow frames, latch intact & \rr release \tabularnewline
    Nut Thread & \rr bolt position & \rr center within 2.5\,mm, height $-2$ to $+1$\,mm, tilt, release, clearance & -- \tabularnewline
    \bottomrule
  \end{tabular}
\end{table*}

\paragraph{\texorpdfstring{$\pi_{0.5}$}{pi0.5}.}
We fine-tune $\pi_{0.5}$~\cite{pi05} with unmodified openpi from \code{pi05\_libero}, changed so that the discretized state enters the prompt, the action includes stiffness, and one model at batch 32 serves all 7 tasks, each with one short fixed prompt without numbers.
With force, the 6 force dims join the state and enter the prompt as tokens of 256 levels, as in the Prompt Fusion of ForceBench~\cite{han2026forcebench}, which used it on other tasks and actions.
About 1 to 2\% of force frames fall outside the normalization range and take the limit tokens, as in openpi.

\paragraph{ForceVLA.}
The other learned rows concatenate force, so we added the force-aware ForceVLA~\cite{yu2025forcevla} lest our conclusion on trained force reading rest on concatenation alone, keeping its model code but changing the tasks, the data, and the steps.
ForceVLA also differs from $\pi_{0.5}$ in backbone, state input, normalization, and batch size, so we do not compare these two rows directly.

\subsection{Code-as-Policies}
\label{sec:supp-caprecipe}

In the program rows of Sec.~5.4 of the main paper, GPT-6 Astra at low reasoning effort writes one Python program per episode in a single call without examples, and the program runs with no further model calls.
Its API comes from the harness tool schemas with the same 25\,N stop and returns kinematics, and our force API adds the force primitives, whose results carry contact events, peak and final forces, and the physics translation layer.
The force row thus measures a single zero-shot program written over our force interface, closer in spirit to code generation over compliant primitives~\cite{burns2024genchip,monwilliams2025ellmer} than to the original Code-as-Policies~\cite{liang2023cap}, which uses few-shot examples, hierarchical generation, and no force.

\section{ForceBench port}
\label{sec:supp-forcebench}

This section gives the details behind Sec.~5.2 and Table~4 of the main paper, all in simulation on the public ForceBench repository~\cite{han2026forcebench} at commit \code{31acf73} with Isaac Sim 6.0.1 and Isaac Lab 3.0.0b2.post1.

\begin{table*}[t]
  \fixedtable{H3}
  \caption{All ten trained methods of the ForceBench paper, success in percent as reported in its Table~I and not reproduced by us, with 200 demonstrations and 200 evaluation episodes per task. Table~4 of the main paper shows six of these rows.}
  \label{tab:supp-fb-trained}
  \addtocounter{table}{-1}
  \centering
  \footnotesize
  \setlength{\tabcolsep}{5pt}
  \begin{tabular}{@{}l cccccccc c@{}}
    \toprule
    Method & Weight & CoM & Faucet & Faucet dir. & Stove & Stove press & Drawer & Nut & Mean \\
    \midrule
    $\pi_{0.5}$, no force & 54.0 & 45.5 & 9.5 & 31.5 & 8.5 & 1.0 & 4.5 & 31.5 & 23.25 \\
    $\pi_{0.5}$ + Token Fusion & 50.5 & 50.0 & 12.0 & 28.5 & 6.0 & 2.5 & 13.0 & 26.0 & 23.56 \\
    $\pi_{0.5}$ + Prompt Fusion & 88.0 & 56.5 & 21.0 & 41.5 & 16.0 & 2.5 & 23.0 & 34.0 & 35.31 \\
    $\pi_{0.5}$ + Cross-Attention & 54.5 & 51.0 & 6.5 & 33.0 & 5.5 & 2.5 & 6.5 & 17.5 & 22.13 \\
    $\pi_{0.5}$ + Token Fusion + Pred & 78.0 & 48.5 & 13.5 & 32.5 & 6.0 & 1.0 & 4.0 & 26.5 & 26.25 \\
    $\pi_{0.5}$ + Temporal Prompt & 90.0 & 69.5 & 17.5 & 31.0 & 7.0 & 21.5 & 2.5 & 22.0 & 32.63 \\
    $\pi_{0.5}$ + Dual-Path Fusion & 99.5 & 72.0 & 22.0 & 30.0 & 17.5 & 19.0 & 4.5 & 16.5 & 35.13 \\
    $\pi_{0}$ & 61.5 & 39.0 & 3.5 & 27.0 & 7.0 & 5.0 & 7.5 & 36.0 & 23.31 \\
    ForceVLA & 35.5 & 45.5 & 12.5 & 14.0 & 10.0 & 12.0 & 20.0 & 26.0 & 21.94 \\
    ForceFlow & 51.5 & 3.0 & 0.0 & 0.5 & 0.0 & 0.0 & 0.0 & 0.0 & 6.88 \\
    \bottomrule
  \end{tabular}
\end{table*}

\begin{table*}[t]
  \fixedtable{I1}
  \caption{The six real-robot tasks as in the final runs. Actions gives the largest number of tool calls in one episode.}
  \label{tab:supp-tasks}
  \addtocounter{table}{-1}
  \centering
  \footnotesize
  \setlength{\tabcolsep}{4pt}
  \let\rr\raggedright
  \begin{tabular}{@{}p{0.17\linewidth}p{0.29\linewidth}p{0.44\linewidth}r@{}}
    \toprule
    \rr Task & \rr Scene & \rr What the hand must do & Actions \tabularnewline
    \midrule
    \rr Wipe a whiteboard & \rr marker writing, an eraser & \rr find the board by touch and wipe until the images show no writing & 40 \tabularnewline
    \rr Drive a screw & \rr a screw and a gear with a threaded center hole, on a wooden board & \rr find the hole by touch, press the tip in, and turn with the thread, grasping again at the wrist limit & 60 \tabularnewline
    \rr Peel an eggplant & \rr an eggplant on foam, a clamped peeler & \rr keep the blade in the curved skin while moving along $+x$ & 60 \tabularnewline
    \rr Insert a USB plug & \rr a USB-A plug on a cable, a power strip & \rr reach a port, find it by a light sliding search, and seat the plug & 40 \tabularnewline
    \rr Sort three bottles by weight & \rr look-alike bottles of 294, 204, and 160\,g, a white and a green plate & \rr lift each by its cap, take the heaviest to the white plate and the lightest to the green one & 80 \tabularnewline
    \rr Tell soft from hard & \rr two wrapped objects of one shape & \rr say which is soft without grasping & 15 \tabularnewline
    \bottomrule
  \end{tabular}
\end{table*}

\setcounter{table}{1}%
\subsection{Setup}
\label{sec:supp-fb-integration}
\label{sec:supp-fb-info}
\label{sec:supp-fb-tools}

The port keeps ForceBench's evaluator and success criteria and adapts only the harness.
\paper serves as a websocket policy that returns a target for the tool center point and a gripper value per request, and an adapter gives the primitives the interface of the robosuite backend, so the primitives, the physics translation layer, and the agent loop are not edited.
Each request also carries both camera calibrations, which only \tool{locate} and \tool{locate\_object} use.
The arm controller is softer than in MuJoCo, so the lead cap while seeking contact is 15\,mm instead of 6\,mm, and the adapter compensates the sag of a held load.
An episode ends after ForceBench's 100\,s of simulated time, 30 turns, or 60 tool calls.

Each turn the model sees the main and wrist RGB views at 224 by 224 pixels, with no depth and no object coordinates.
Each instruction adds to ForceBench's task sentence a hand-written paragraph of dimensions, key heights, and facts from the public task code, such as the two mass classes on Weight Sort, but never an object's position or a hidden value.
\paper offers 16 to 19 tools per task and the no-force control 9 to 11, and on Stove Press and Nut Thread the control's \tool{move\_turn} takes the place of \tool{press\_turn}, while Nut Thread offers \tool{move\_turn} under \paper too.
Tools and perception thresholds were developed on seeds 300 to 307, and after first runs on seeds 300 to 309 Nut Thread gained \tool{locate\_object} with a corrected instruction and \tool{press\_turn} came to print the resisting torque, first as a mean and then per side.

\begin{table}[t]
  \caption{Successes out of 10 per task on seeds 300 to 309, with \paper as Force, the no-force control as None, and Pairs counting the seeds on which only \paper succeeds, then only the control. On Stove Press, \paper uses the mean report for Astra and the per-side report for Sol, and Nut Thread uses its revised version in all four conditions.}
  \label{tab:supp-fb-results}
  \label{tab:supp-fb-paired}
  \centering
  \footnotesize
  \setlength{\tabcolsep}{3.5pt}
  \begin{tabular}{@{}l ccc ccc@{}}
    \toprule
    & \multicolumn{3}{c}{GPT-6 Astra} & \multicolumn{3}{c}{GPT-6.1 Sol} \\
    \cmidrule(lr){2-4}\cmidrule(l){5-7}
    Task & Force & None & Pairs & Force & None & Pairs \\
    \midrule
    Weight Sort & 10 & 8 & 2 / 0 & 10 & 9 & 1 / 0 \\
    CoM Sort & 7 & 5 & 3 / 1 & 6 & 5 & 4 / 3 \\
    Faucet Turn & 10 & 10 & 0 / 0 & 10 & 9 & 1 / 0 \\
    Faucet Direction & 5 & 5 & 4 / 4 & 4 & 3 & 4 / 3 \\
    Stove Turn & 10 & 10 & 0 / 0 & 10 & 10 & 0 / 0 \\
    Stove Press & 9 & 8 & 1 / 0 & 10 & 6 & 4 / 0 \\
    Drawer Pull & 10 & 0 & 10 / 0 & 10 & 0 & 10 / 0 \\
    Nut Thread & 9 & 10 & 0 / 1 & 10 & 10 & 0 / 0 \\
    \midrule
    All tasks & 70/80 & 56/80 & 20 / 6 & 70/80 & 52/80 & 24 / 6 \\
    \bottomrule
  \end{tabular}
\end{table}

\subsection{Tasks and results}
\label{sec:supp-fb-tasks}
\label{sec:supp-fb-scripted}
\label{sec:supp-fb-results}
\label{sec:supp-fb-cost}

\Cref{tab:supp-fb-tasks} lists what each task hides and how the public evaluator decides success.
On Faucet Turn, Stove Turn, and Drawer Pull it omits conditions that the ForceBench paper lists, shown in the last column.
A scripted state machine per task on our tools succeeds in 50/64 episodes on seeds 300 to 307 and in 48/64 on seeds 308 to 315, which shows that the tasks are feasible from RGB.

\Cref{tab:supp-fb-results} gives the per-task counts behind Table~4 of the main paper, and \cref{tab:supp-fb-trained} lists all ten trained methods of the ForceBench paper, which learn from 200 demonstrations per task, see only the task sentence, and are evaluated on seeds 300 to 499.
The force gain is largest on Drawer Pull.
In 19 of 20 episodes with force, the model calls \tool{pull} once with a 2.4\,N target, and the hand stops where the resistance balances it, 1.1 to 1.7\,mm from the goal against a 2.5\,mm window.
Without force no fixed motion can work, since the target opening ranges from 63.5 to 90.7\,mm, and all 20 episodes stop short.

\section{Real-robot system and tasks}
\label{sec:supp-real}

This section summarizes how the harness runs on the real robot of Sec.~5.1 of the main paper, a Franka FR3 without a wrist force sensor, and every constant here belongs to that arm.

\paragraph{System.}
\label{sec:supp-real-system}
The stock joint-impedance controller of Polymetis runs unchanged at 1\,kHz, with joint gains at four times the default and eight times on joint 7, and we move only its joint target.
Beside it on the real-time computer, our guard runs at a nominal 50\,Hz and enforces the hard limits of \cref{sec:supp-prim-constants}.
The harness on the workstation runs the primitives and estimators and sends the guard one step per tick, at a median rate of 14.9\,Hz.
The frozen model is called once per turn and answers with one tool call.

\paragraph{Force estimate.}
\label{sec:supp-sensing}
Polymetis forwards the filtered external joint torques that libfranka estimates but not a wrench, so the guard solves Eq.~1 of the main paper on every tick by least squares with the flange Jacobian of the controller's own kinematic model.
It also checks the joint torque that no flange wrench explains, and a residual above 3\,N$\cdot$m or a Jacobian near a singularity ends the episode with a fault.
The estimate has little noise, 0.03 to 0.05\,N at rest, but its zero shifts by a few newtons with the approach direction, joint friction in motion, and the wrist angle.
Filtering cannot remove such an offset, so we calibrate only the offset and read force by the three rules of Sec.~3.2 of the main paper.
The harness tares with the gripper open before each empty-hand primitive, and after every primitive it relaxes the target onto the measured pose for 0.8\,s and then averages the wrench over 0.5\,s.
Relaxing lowers the median difference between consecutive still readings from 1.07\,N to 0.08\,N over 999 empty-hand windows, because a target that leads the joints pushes against joint friction, which the estimate reports as external force.
Each still reading is reported relative to the same reading at the start of the episode, and with an object in the hand also as the change since the grasp.
Lifted weights are therefore compared within an episode rather than read in grams.

\paragraph{Safety.}
Beneath the guard of \cref{sec:supp-prim-constants}, the harness adds collision stops at 8\,N, or 12\,N when holding, in the guarded tools of \cref{sec:supp-primitives}, and it ends an episode after the number of actions in \cref{tab:supp-tasks}.
An operator stands at the emergency stop in every episode.

\paragraph{Localization.}
\label{sec:supp-loc}
The exterior camera is uncalibrated, so every coordinate comes from fixed geometry on the wrist depth image, with no learned detector.
A pixel maps to the base frame through the flange pose at capture and the hand--eye calibration of the wrist camera.
Each turn, the harness groups points about 1\,cm or more above the fitted table plane into objects and lists their center, top, and long axis with its size, and the model picks targets itself with \tool{locate}.
The hand--eye calibration has a median error of 4.1\,mm and 0.28$^\circ$ on 6 held-out views.
The model is told that coordinates are good to about 1 to 2\,cm, so vision alone cannot reach a port, and its last millimeters are left to \tool{search}.

\paragraph{Tasks and conditions.}
\Cref{tab:supp-tasks} describes each task as in the final runs, and Sec.~5.1 of the main paper gives the success criteria.
\paper offers all 22 tools of \cref{sec:supp-primitives} except \tool{wiggle} and \tool{press\_turn}, USB insertion adds \tool{wiggle}, and screw driving swaps \tool{lift} for \tool{wiggle}.
The no-force control keeps the 11 tools that move by position only or not at all, and its tool results report only kinematics.
In peeling it also gets \tool{move\_until\_contact}, which stops at a fixed 5\,N and reports no force.

\section{Limitations and reproducibility}
\label{sec:supp-limits-repro}

\subsection{Limitations}
\label{sec:supp-validity}

Querying a remote frontier model adds seconds of latency between primitives, which suits quasi-static contact tasks but precludes fast reactive replanning during dynamic motion.
Both the primitive menu and the physics estimators are specified by hand, and each task relies on a hand-written instruction, leaving automatic primitive synthesis and open-ended task specification to future work.
Finally, our evaluation focuses on single-arm tabletop manipulation with a parallel gripper across models from two API providers, and our component ablations are conducted in simulation rather than repeated on real hardware.

\subsection{Reproducibility}
\label{sec:supp-repro}

\bench uses robosuite 1.5.1 and MuJoCo 3.2.7, ForceBench runs on Isaac Lab with the changes of \cref{sec:supp-fb-integration}, and the real robot runs stock Polymetis with its hybrid joint impedance controller.
Code and configurations will be released upon publication, together with logs and task assets.

Model settings are those of Sec.~5 of the main paper, and the GPT-5.x models use chat completions with reasoning off when tools are present.

\end{document}